\documentclass{article}

\usepackage{stackengine}

\usepackage[accepted]{icml2020}
\usepackage{hyperref,soul}

\usepackage[utf8]{inputenc} 
\usepackage[T1]{fontenc}    

\usepackage{hyperref}       
\usepackage{url}            
\usepackage{booktabs}       
\usepackage{amsfonts,amsmath}       
\usepackage{nicefrac}       
\usepackage{microtype}      
\DeclareMathOperator*{\argmax}{arg\,max}
\DeclareMathOperator*{\argmin}{arg\,min}
\usepackage{graphicx}
\usepackage{support-caption}  
\usepackage{subcaption}
\usepackage{mathtools}
\usepackage{tabularx}
\usepackage{float}
\usepackage{stfloats}
\usepackage{cancel}

\usepackage{wrapfig}
\usepackage{bm}
\usepackage{algorithm}
\usepackage[]{algorithmic}
\usepackage[title]{appendix}
\usepackage{enumitem}
\usepackage[dvipsnames]{xcolor}

\icmltitlerunning{A Model-Based Derivative-Free Approach to Black-Box Adversarial Examples: BOBYQA}

\begin{document}

\twocolumn[
\icmltitle{A Model-Based Derivative-Free Approach to Black-Box Adversarial Examples: BOBYQA}



\icmlsetsymbol{equal}{*}

\begin{icmlauthorlist}
\icmlauthor{Ughi Giuseppe}{ed}
\icmlauthor{Abrol Vinayak}{ed}
\icmlauthor{Tanner Jared}{ed}
\end{icmlauthorlist}

\icmlaffiliation{ed}{Mathematical Institute, University of Oxford, Oxford, United Kingdom}

\icmlcorrespondingauthor{Ughi Giuseppe}{ughi@maths.ox.ac.uk}

\icmlkeywords{Machine Learning, ICML, Adversarial Examples, DFO}

\vskip 0.3in]



\printAffiliationsAndNotice{}  

\begin{abstract}

We demonstrate that model-based derivative free optimisation algorithms can generate adversarial targeted misclassification of deep networks using fewer network queries than non-model-based methods.  Specifically, we consider the black-box setting,  and show that the number of networks queries is less impacted by making the task more challenging either through reducing the allowed $\ell^{\infty}$ perturbation energy or training the network  with defences against  adversarial misclassification.  We illustrate this by contrasting the BOBYQA algorithm \cite{powellbobyqa} with the state-of-the-art model-free adversarial targeted misclassification approaches based on genetic \cite{Alzantot}, combinatorial \cite{COMBI}, and direct-search \cite{andriushchenko2019square}  algorithms.  We observe that for high $\ell^{\infty}$ energy perturbations on networks, the aforementioned simpler model-free methods require the fewest queries. In contrast, the proposed BOBYQA based method achieves state-of-the-art results when the perturbation energy decreases, or if the network is trained against adversarial perturbations.

\end{abstract}

\section{Introduction}

Deep neural networks (NNs) achieve state-of-the-art performance on a growing number of applications such as acoustic modelling, image classification, and fake news detection \cite{hinton2012deep,he,monti2019fake} to name but a few.  Alongside their growing application, there is a literature on the robustness of deep nets which shows that it is often possible to generate images with subtle perturbations, referred to as adversarial examples \cite{szegedy2013, goodfellow2014}, to the input of a network resulting in its performance being severely degraded; for example, see \cite{dalvi,kurakin,sitawarin,  eykholt2017robust,yuan} concerning the use-case of self driving cars.  

Methods to generate these adversarial examples are classified according to two main criteria \cite{yuan}.
\begin{description}
    \item[Adversarial Specificity] establishes what the aim of the adversary is. In \textit{non-targeted}  attacks, the method perturbs the image in such a way that it is misclassified into any different category than the original one. While in \textit{targeted} settings, the adversary specifies a category into which an image has to be misclassified.

     \item[Adversary's Knowledge] defines the amount of information available to the adversary. In \textit{White-box} settings the adversary has complete knowledge of  the network architecture and weights, while in the \textit{Black-box} setting the adversary is only able to obtain the pre-classification outpupt vector for a limited number of inputs.  The White-box setting allows for the use of gradients of a missclassification objective to efficiently compute the adversarial example \cite{goodfellow2014,carlini,Chen2018ead},  while the  same optimization formulation of the Black-box setting requires use of a derivative free approach \cite{narodytska2017,chen,ilyas2018black,Alzantot}.
     
    
\end{description}

\begin{figure}[th!]
\centering
\includegraphics[width=.8\linewidth]{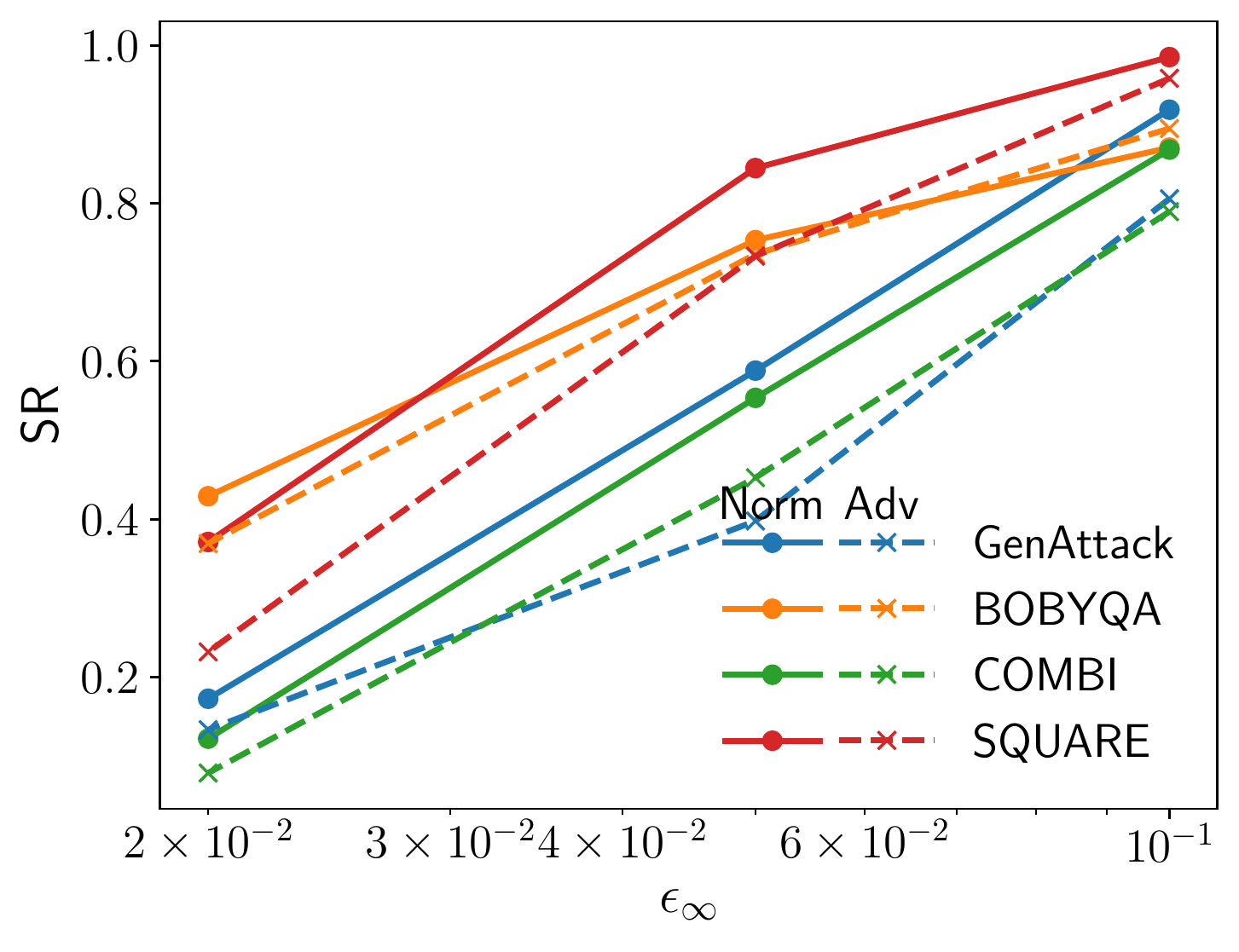}
\caption{The success rate (SR) of the BOBYQA algorithm to generate a targeted adversarial example compared to GenAttack \cite{Alzantot}, COMBI \cite{COMBI}, and SQUARE \cite{andriushchenko2019square} attacks as a function of the perturbation energy; specifically for a network trained on the CIFAR10 dataset without defences (Norm) and with the distillation defence (Adv) \cite{Papernot_distillation}. It can be observed that as $\varepsilon_\infty$ decreases the BOBYQA based method achieves a higher SR than other methods. Similarly, the success rate of BOBYQA is less affected by adversarial training. In particular, with the infinity norm of the perturbation limited to ${\varepsilon_\infty=.02}$ BOBYQA achieves a SR 1.15 and 1.59 folds better than SQUARE when considering Norm and Adv respectively. Here the number of  network queries were restricted to 3,000, for further details see Fig. \ref{fig:CIFARcdf}.}

\label{fig:compa}
\end{figure}

The generation of black-box targeted adversarial examples for deep NNs has been extensively studied in a setting initially proposed by \cite{chen} where:
\begin{itemize}
    \item the adversarial example is found by solving an optimisation problem designed to change the original classification of a specific input to a specific alternative.
    \item  the perturbation, which causes the network to change the classification, has entries bounded in magnitude by a specified infinity norm (maximum entry magnitude).
    \item the \textit{number of queries} to the NN needed to generate the adversarial example should be  as small as possible.
\end{itemize}

The Zeroth-Order-Optimization (ZOO) \cite{chen} introduced DFO methods for computing adversarial examples in the black-box setting, specifically using a coordinate descent optimization algorithm.   At the time this was a substantial departure from methods for the black-box setting which train a proxy NN and then employ gradient based methods for white-box attacks on the proxy network \cite{papernot, tu2018}; such methods are especially effective when numerous adversarial examples will be computed, but require substantially more network queries than the methods designed for misclassifying individual examples.    Following the introduction of ZOO, there have been numerous improvements using other model-free DFO based approaches, see  \cite{Alzantot,COMBI, andriushchenko2019square}.   Specifically, GenAttack \cite{Alzantot} is a genetic algorithm, COMBI \cite{COMBI} is a direct-search method  that explores the vertices of  the perturbation energy, and SQUARE \cite{andriushchenko2019square} is a randomized direct-search method.

In this manuscript we consider an alternative model-based DFO method based on BOBYQA \cite{powellbobyqa} which explicitly develops \textit{models that approximate the loss function} in the optimisation problem and minimises the models using methods from continuous optimisation. By considering adversarial perturbations to three NNs trained on different datasets (MNIST, CIFAR10, and ImageNet), we show that for the model-free methods \cite{Alzantot, COMBI, andriushchenko2019square} the number of evaluation of the NN grows more rapidly as the maximum perturbation energy decreases than does the method built upon BOBYQA. As a consequence GenAttack, COMBI and SQUARE are preferable for large values of the maximum perturbation energy and BOBYQA for smaller values. As an example Figure \ref{fig:compa} illustrates how the BOBYQA based algorithm compares to GenAttack, COMBI, and SQUARE when considering a net either normally or adversarially trained on CIFAR10 with different maximum perturbation energies.

We observe the intuitive principle that direct-search methods are effective to misclassify 
NNs with high perturbation energies, while in more challenging settings it is preferable to use more sophisticated model-based methods, like ours.  Model-based approaches will further challenge defences to adversarial missclassification \cite{dhillon2018stochastic,ijcai2019-833}, and in so doing will lead to improved defences and more robust networks.
Model-based DFO is a well developed area, and we expect further improvements are possible through a more extensive investigation of these approaches.


\section{Adversarial Examples Formulated as an Optimisation Problem}\label{sec:opt}

Consider a classification operator $F:\mathcal{X}\rightarrow\mathcal{C}$ from input space $\mathcal{X}$ to output space $\mathcal{C}$ of classes. A targeted adversarial perturbation $\bm{\eta}$ to an input $\textbf{X} \in \mathcal{X}$ has the property that it changes the classification to a specified target class $t$, i.e $F(\textbf{X})= c$ and $ F(\textbf{X}+\bm{\eta})=t \neq c$.
Herein we follow the formulation by \cite{Alzantot}. Given: an image $\textbf{X}$, a maximum energy budget $\varepsilon_{\infty}$, and a suitable loss function $\mathcal{L}$, then the task of computing the adversarial perturbation $\bm{\eta}$ can be cast as an optimisation problem such as
\begin{align}
    &\min_{\bm{\eta}} \; \mathcal{L}(\textbf{X},\bm{\eta})\;
    s.t.\; \|\bm{\eta} \|_\infty \leq \varepsilon_\infty \label{eq:opt_prob_beg}\\
    & [\textbf{X} + \bm{\eta}]_j \geq l;\; [\textbf{X} + \bm{\eta}]_j \leq u \;\; \forall j \in {1,\ldots,n}, \nonumber \label{eq:opt_prob_end}
\end{align}
where the final two inequality constraints are due to the input entries being restricted to $[l,u]^n$.  Denoting the pre-classification output vector by $f(\textbf{X})$, i.e. ${F(\textbf{X}) = \argmax f(\textbf{X})}$, then the misclassification of $\textbf{X}$ to target label $t$ is achieved by $\bm{\eta}$ if $f(\textbf{X} + \bm{\eta})_t \geq \max_{j\neq t} f(\textbf{X} + \bm{\eta})_j$. In \cite{carlini,chen,Alzantot} they determined   
\begin{equation}\label{eq:loss}
    \mathcal{L}(\textbf{X},\bm{\eta}) = \log\left(\Sigma_{j\neq t}f(\textbf{X}+\bm{\eta})_j\right) - \log\left(f(\textbf{X}+\bm{\eta})_t\right),
\end{equation}
to be the most effective loss function for computing $\bm{\eta}$ in \eqref{eq:opt_prob_beg}, and we also employ this choice throughout our experiments.

\section{Derivative Free Optimisation for Adversarial Examples}

Derivative Free Optimisation is a well developed field with numerous types of algorithms, see \cite{conn2009introduction} and \cite{larson_menickelly_wild_2019} for reviews on DFO principles and algorithms.  Examples of classes of such methods include: direct search methods such as simplex, model-based methods, hybrid methods such as finite differences or implicit filtering, as well as randomized variants of the aforementioned and methods specific to convex or noisy objectives.  The optimization formulation in Section~\ref{sec:opt} is amenable to virtually all DFO methods, making it unclear which of the algorithms to employ.  Methods which have been trialled include: the finite difference based ZOO attack \cite{chen}, a combinatorial direct search of the perturbation $\ell^{\infty}$ energy constraint method COMBI \cite{COMBI}, a genetic direct search method GenAttack \cite{Alzantot}, and most recently a randomized direct-search method \cite{andriushchenko2019square}.  Notably missing from the aforementioned list are model-based methods.

Given a set of $q$ samples $\mathcal{Y} = \{\textbf{y}^1,...,\textbf{y}^q\}$ with $\textbf{y}^i$ $\in \mathbb{R}^n$, model-based DFO methods start by identifying the minimiser of the objective among the samples at iteration $k$, $\textbf{x}^k =\argmin_{\textbf{y}\in \mathcal{Y}} \mathcal{L}(\textbf{y})$. Following this, a model for the objective function $\mathcal{L}$ is constructed, typically centred around the minimizer. In its simplest form one uses a polynomial approximation to the objective,  such as a quadratic model centred in $\textbf{x}^k$
\begin{equation}\label{eq:Model}
    m(\textbf{x}^k + \textbf{p}) = a + \textbf{c}^\top \textbf{p} + \frac{1}{2}\textbf{p}^\top \textbf{M} \textbf{p},
\end{equation}
with $a\in\mathbb{R}$, $\textbf{c}$, $\textbf{p}\in\mathbb{R}^n$, and $\textbf{M}\in \mathbb{R}^{n\times n}$ being also symmetric. In a white-box setting one would set $\textbf{c} = \nabla \mathcal{L}(\textbf{x}^k)$ and $\textbf{M}= \nabla^2 \mathcal{L}(\textbf{x}^k)$, but this is not feasible in the black-box setting as we do not have access to the derivatives of the objective function. Thus $\textbf{c}$ and $\textbf{M}$ are usually defined by imposing interpolation conditions 
\begin{equation}\label{eq:interpolation}
 m_k(\textbf{y}^i) = \mathcal{L}(\textbf{y}^i) \ \ \ \ \text{   for } i = 1,2,...,q,
\end{equation}
and when $q<1 + n + n(n+1)/2$ (i.e. the system of equations is under-determined) other conditions are introduced according to which algorithm is considered. The objective model \eqref{eq:Model} is considered to be a good estimate of the objective in a neighbourhood referred to as a trust region.  Once the model $m_k$ is generated, the update step $\textbf{p}$ is computed by solving the trust region problem
\begin{equation}\label{eq:Hessain}
    \min_\textbf{p} m_k(\textbf{x}_k + \textbf{p}), \ \ \ \text{subject to } \ \| \textbf{p}\| \leq \Delta,
\end{equation} 
where $\Delta$ is the radius of the region where we believe the model to be accurate, for more details see \cite{nocedal}. The new point $\textbf{x}_k + \textbf{p}$ is added to $\mathcal{Y}$ and a prior point is potentially removed. Herein we consider an exemplary\footnote{BOBYQA was selected among the numerous types of model-based DFO algorithms due to its efficiency observed for other similar problems requiring few model samples as in climate modelling \cite{Climate}} model-based method called BOBYQA.

\paragraph{BOBYQA}

The BOBYQA algorithm, introduced in \cite{powellbobyqa}, updates the parameters of the model $a,\textbf{c},$ and $\textbf{M}$, in each iteration in such a way as to minimise the change in the quadratic term $\textbf{M}_k$ between iterates while otherwise fitting the sample values:
\begin{align}
    &\min_{a_k,\textbf{c}_k,\textbf{M}_k} \| \textbf{M}_k - \textbf{M}_{k-1}\|_F^2 \\
    &s.t.   \;   m_k(\textbf{y}^i) = \mathcal{L}(\textbf{y}^i), \  i = 1,2,\ldots,q,
\end{align}
with $n+1 < q <  1 + n + n(n+1)/2$ and $\textbf{M}_k$ initialised as the zero matrix. When the number of parameters $q = n+1$ then the model is considered as linear with $\textbf{M}_k$ set as zero. 
We further allow only $\kappa$ queries at each implementation of BOBYQA, since after the model is generated few iterations are needed to find the minimum. 

\subsection{Computational Scalability and Efficiency}

For improved computational scalability and efficiency, we do not solve \eqref{eq:opt_prob_beg} for $\bm{\eta}\in\mathbb{R}^n$ directly, but instead use domain sub-sampling and hierarchical liftings: domain sub-sampling iteratively sweeps over batches of $b$$<<$$n$ variables, see \eqref{eq:opt_sub_samp}, while hierarchical liftings clusters and perturbs variables simultaneously, see \eqref{eq:opt_sub_samp_fin}. 

\paragraph{Domain Sub-Sampling}

\begin{figure}[t!]
\centering
\includegraphics[width=.85\linewidth]{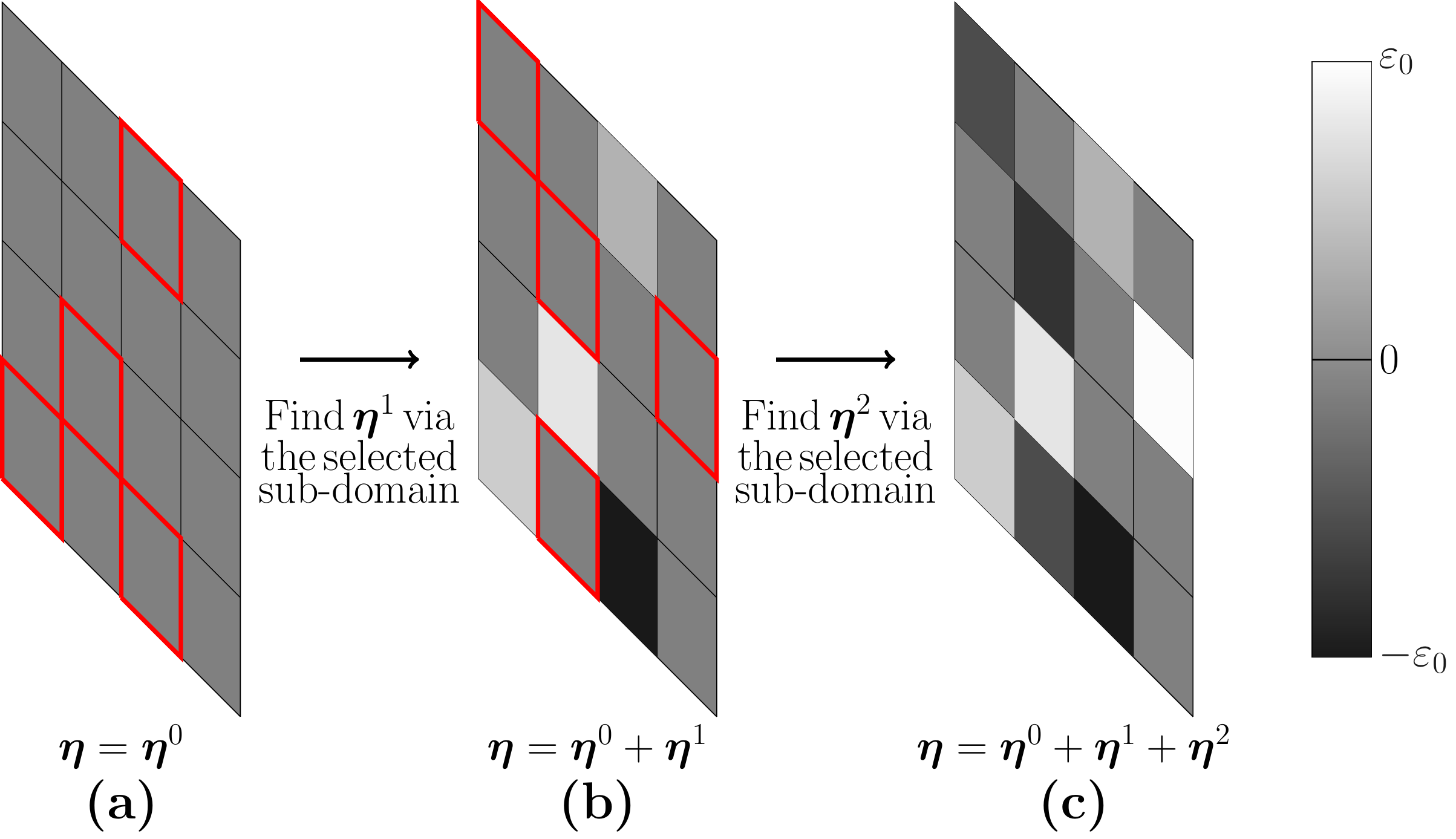}
\caption{Example of how the perturbation $\bm{\eta}$ evolves through the iterations when an image in $\mathbb{R}^{4\times4}$ is attacked. In (a) the perturbation is $\bm{\eta} = \bm{\eta}^0$ and we select a sub-domain of $b=4$ pixels (in red). Once we have found the optimal perturbation $\bm{\eta}^1$ in the selected sub-domain, we update the perturbation in (b) and select a new sub-domain of dimension $b$. The same is repeated in (c).}
\label{fig:Sub-Domain}
\end{figure}

The simplest version of domain sub-sampling consists of partitioning input dimension $n$ into smaller disjoint domains; for  example,  $k= \lfloor n/b \rfloor$ domains $\Omega^j$ of  size $b\ll n$ which are disjoint and which cover all of $[n]$. 
Rather than solving \eqref{eq:opt_prob_beg} for $\bm{\eta}\in\mathbb{R}^n$ directly, for each of  $j=1,\ldots,k$ one {\em sequentially} solves for $\bm{\eta}^j\in\mathbb{R}^n$ which are only non-zero for entries in $\Omega^j$.  The resulting sub-domain perturbations $\bm{\eta}^j$ are then summed to generate the full  perturbation     
$\bm{\eta} = \sum_{j=1}^k \bm{\eta}^j$, see Figure \ref{fig:Sub-Domain} as an example.  That is, the optimisation problem (\ref{eq:opt_prob_beg}) is adapted to repeatedly looping over $j=1,\ldots,k$: 
\begin{align}
    &\min_{\bm{\eta^j}} \;\; \mathcal{L}\left(\textbf{X}+\sum_{h\neq j}\bm{\eta}^{\ell},\bm{\eta}^j\right) 
    \; s.t. \; \left\|\sum_{h=1}^k \bm{\eta}^{h} \right\|_\infty \leq \varepsilon_\infty  \label{eq:opt_sub_samp}\\
    & \left[\textbf{X} +\sum_{h=1}^k\bm{\eta}^{h}  \right]_r \geq l ; \; \left[\textbf{X} + \sum_{h=1}^k\bm{\eta}^{h} \right]_r \leq u \;\; \forall r \in \Omega^j\nonumber 
    \label{eq:opt_prob_proj_end}
\end{align}
where the $\Omega^j$ may be reinitialised; in particular following each loop over $\mathbb{R}^n$ which occurs at $j=k$.

\begin{figure}[t!]
\centering
\rotatebox{90}{\parbox{2mm}{\textbf{MNIST}}}
\begin{subfigure}{.47\columnwidth}
  \centering
  \includegraphics[width=.99\linewidth]{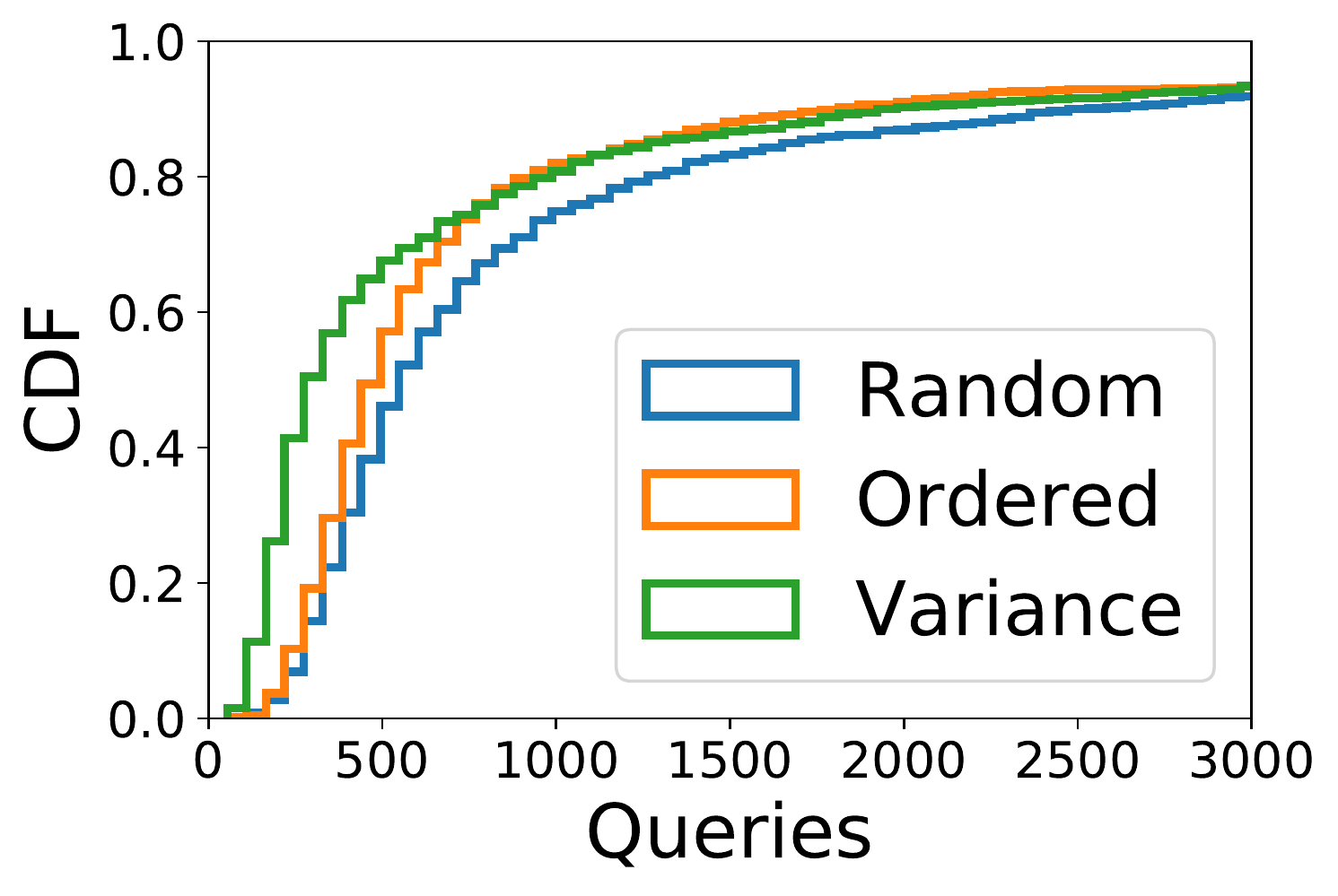}
  \caption{$\varepsilon_\infty = 0.4$}
\end{subfigure}%
\begin{subfigure}{.47\columnwidth}
  \centering
  \includegraphics[width=.99\linewidth]{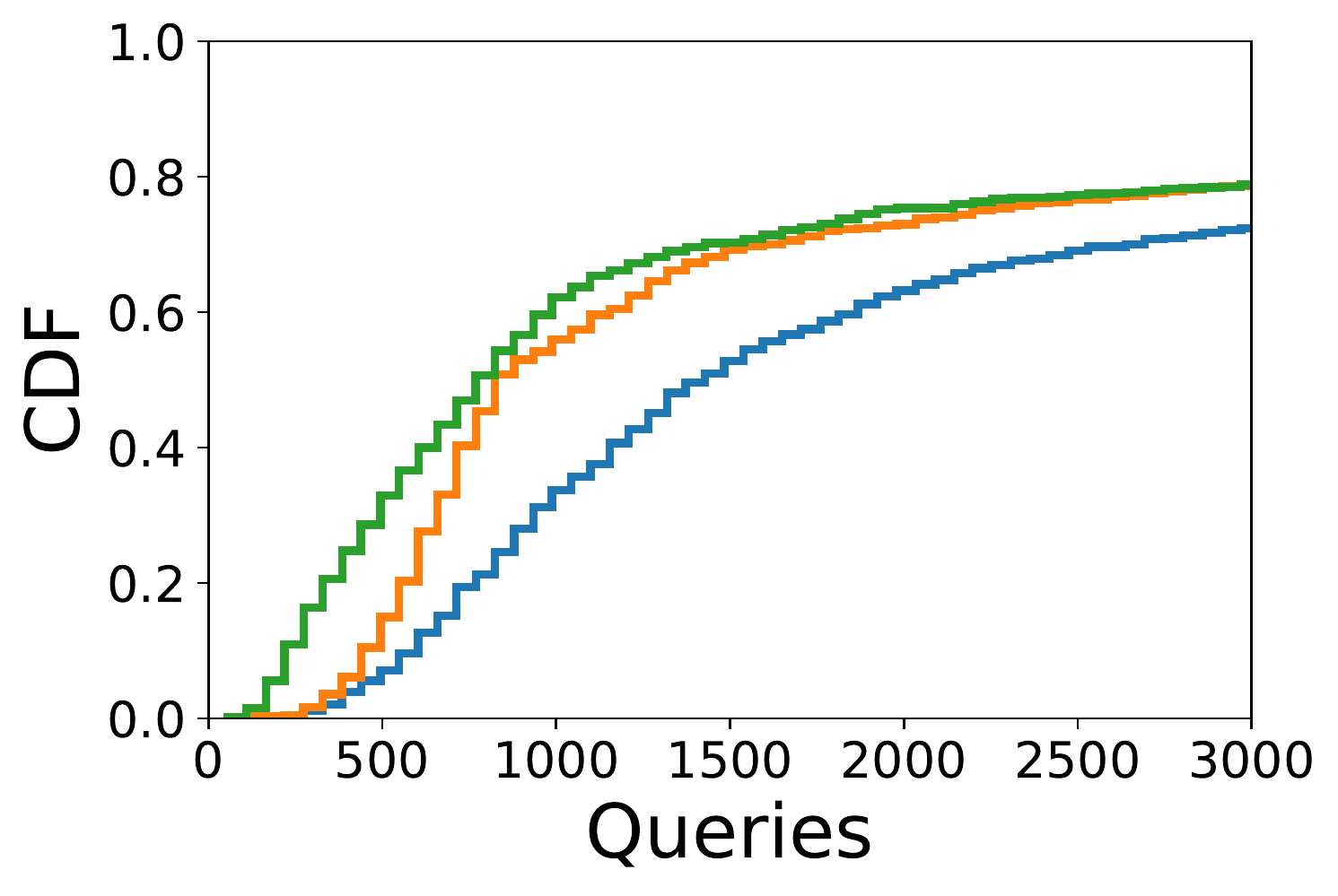}
  \caption{$\varepsilon_\infty = 0.2$}
\end{subfigure}\\
\rotatebox{90}{\parbox{2mm}{\textbf{CIFAR10}}}
\begin{subfigure}{.47\columnwidth}
  \centering
  \includegraphics[width=.99\linewidth]{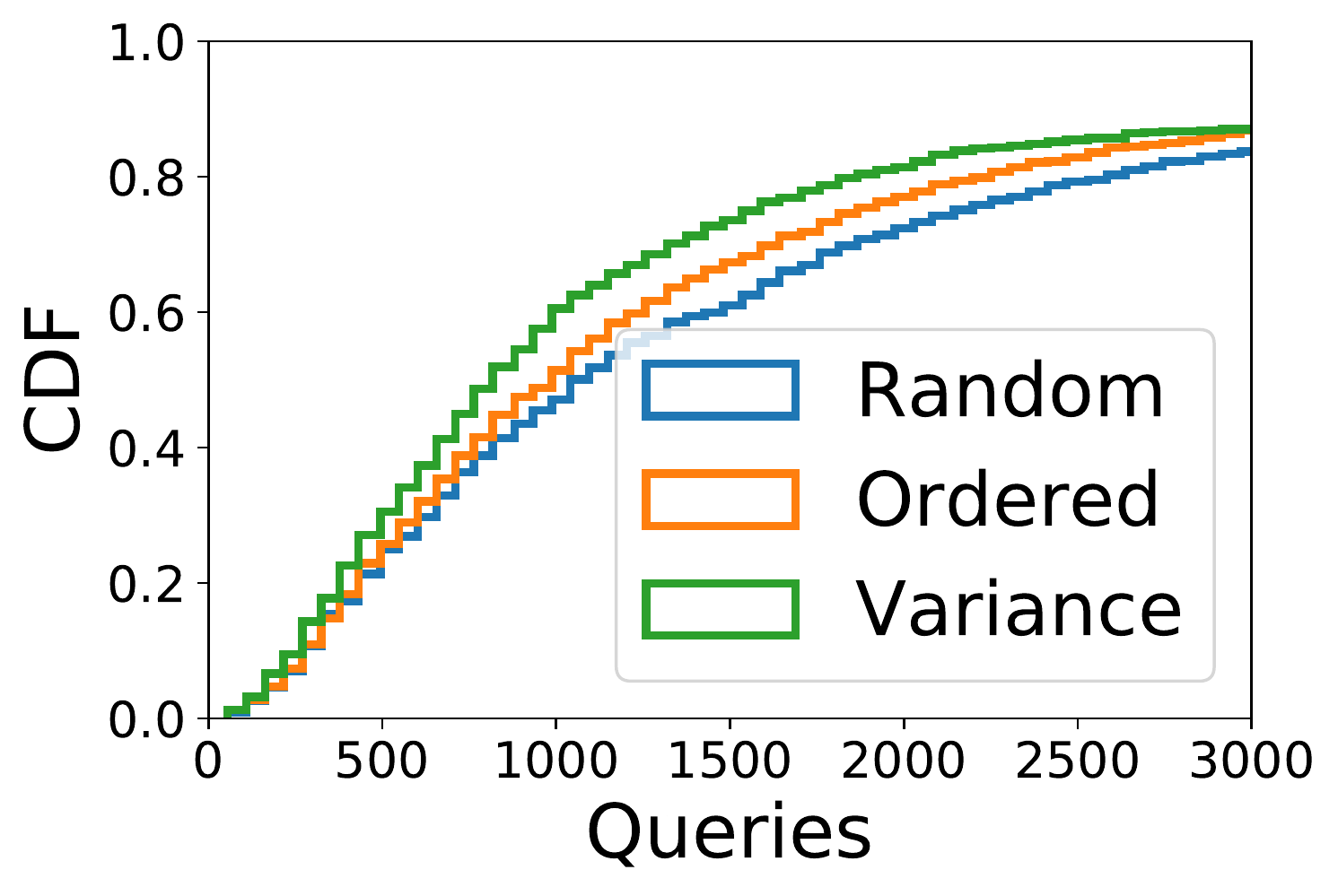}
  \caption{$\varepsilon_\infty = 0.1$}
\end{subfigure}%
\begin{subfigure}{.47\columnwidth}
  \centering
  \includegraphics[width=.99\linewidth]{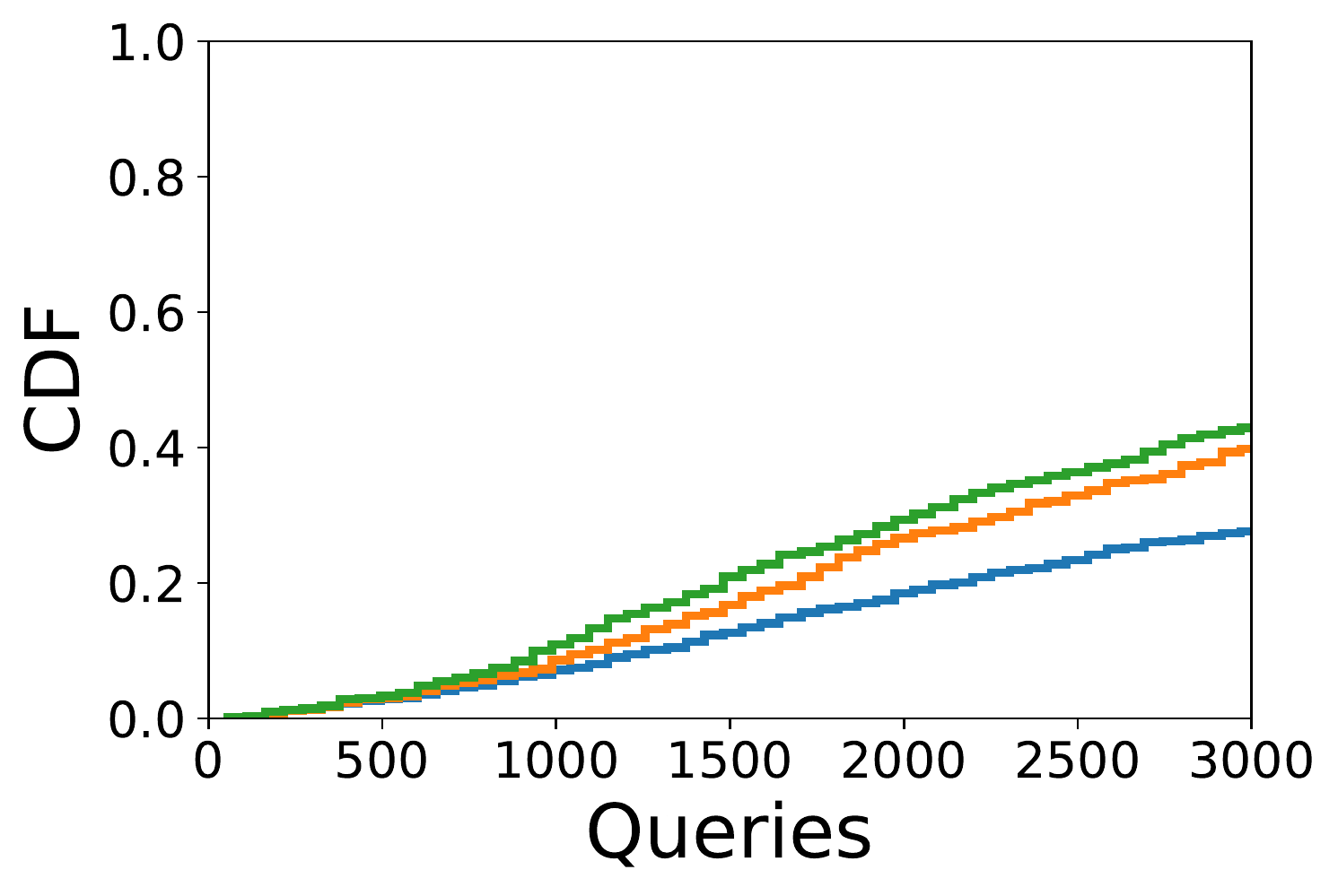}
  \caption{$\varepsilon_\infty = 0.02$}
\end{subfigure}%
\caption{Cumulative distribution function of successfully perturbed images as a function of number of queries to a NN trained on MNIST and CIFAR10 datasets. In each image the effectiveness of different sub-sampling methods in generating a successful adversarial example is shown for different values of perturbation energies $\varepsilon_\infty$. See Section \ref{sec:exp} for details about experimental setup and NN architectures.}
\label{fig:subsamp}
\vskip -0.2in
\end{figure}

We considered three possible ways of selecting the domains 
\begin{itemize}
    \item In \textit{Random Sampling} we consider at each iteration a different random sub-samplings of the domain, $\Omega^1$.
    \item In \textit{Ordered Sampling} we generate a random disjoint partitioning of the domain. Once each variable has been optimised over once a new partitioning is generated.
    \item In \textit{Variance Sampling} we choose $\Omega^j$ to select in decreasing order of local variance of $X$, the variance in intensity among the 8 neighbouring variables (e.g. pixels) in the same colour channel. We further reinitialise $\Omega^j$ after each loop through $j=1,\ldots,k$.
\end{itemize}

In Figure \ref{fig:subsamp} we compare how these different sub-sampling techniques perform when generating adversarial example for the MNIST and CIFAR10 dataset.  It can be observed that variance sampling consistently performs better than random and ordered sampling. This suggest that pixels belonging to high-contrast regions are more influential than the ones in a low-contrast one, and hence variance sampling is the preferable ordering.

\begin{figure}[t!]
\centering
\includegraphics[scale=.4]{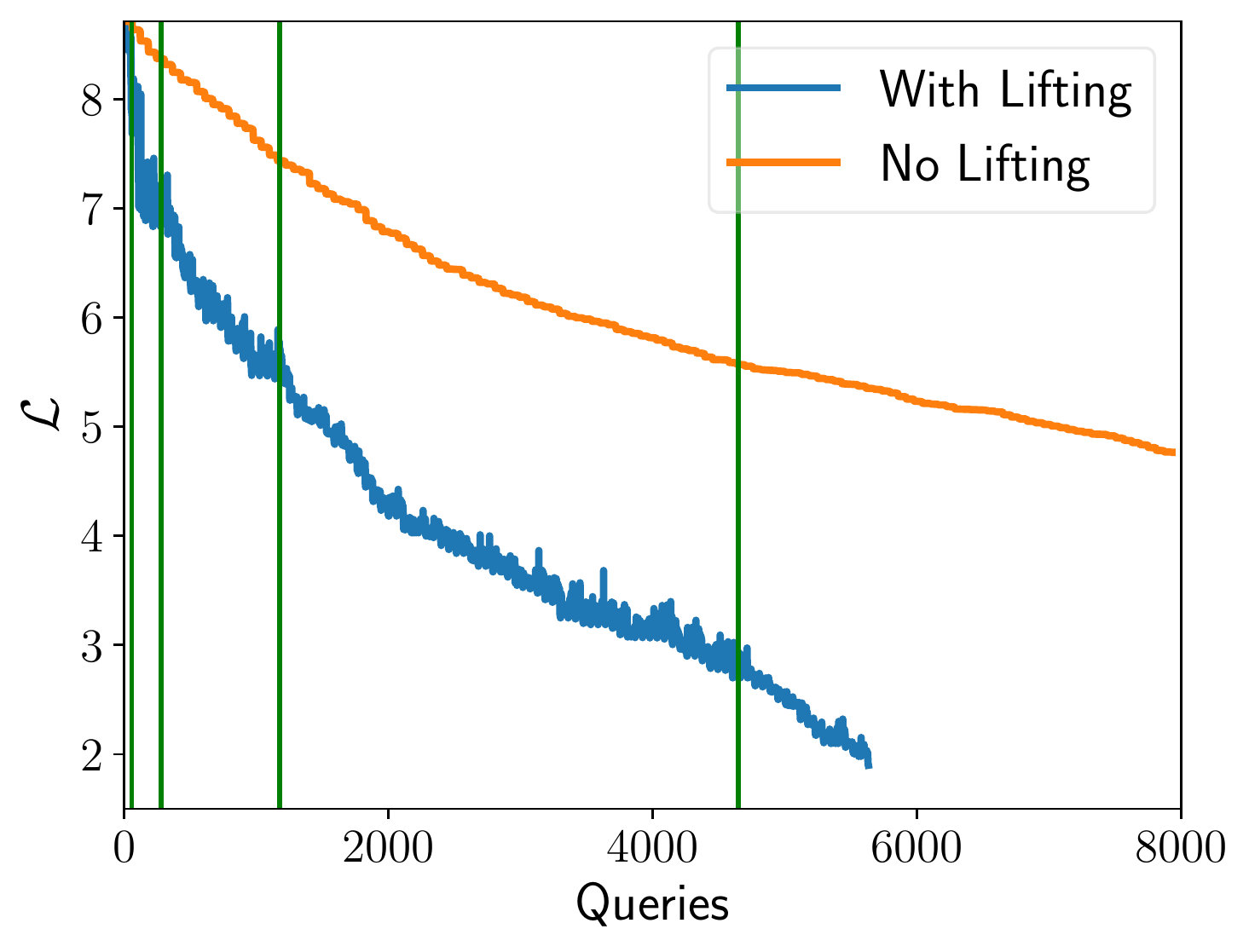}
\caption{Impact of hierarchical lifting approach on Loss function \eqref{eq:loss} as a function of the number of queries to Inception-v3 net trained on ImageNet dataset to find the adversarial example for a single image. The green vertical lines correspond to changes of hierarchical level, which entail an increase in the dimension of the optimisation space.}
\label{fig:resize2}
\end{figure}

\paragraph{Hierarchical Lifting}

When the domain is very high dimensional, working on single pixels is not efficient as the above described method would imply modifying only a very small proportion of the image; for instance, we will choose $b=50$ even when $n$ is almost three-hundred-thousand. Thus to perturb wider portions of the image, we consider a hierarchy of liftings as in the ZOO attack presented in \cite{chen}. We seek an adversarial example by optimising over increasingly higher dimensional spaces at each step referred here as level $\ell$ lifted to the image space. As an illustration, Figure \ref{fig:resize2} shows that hierarchical lifting has a significant impact on the minimisation of the loss function.

\begin{figure}[t!]
\centering
\includegraphics[scale=.3]{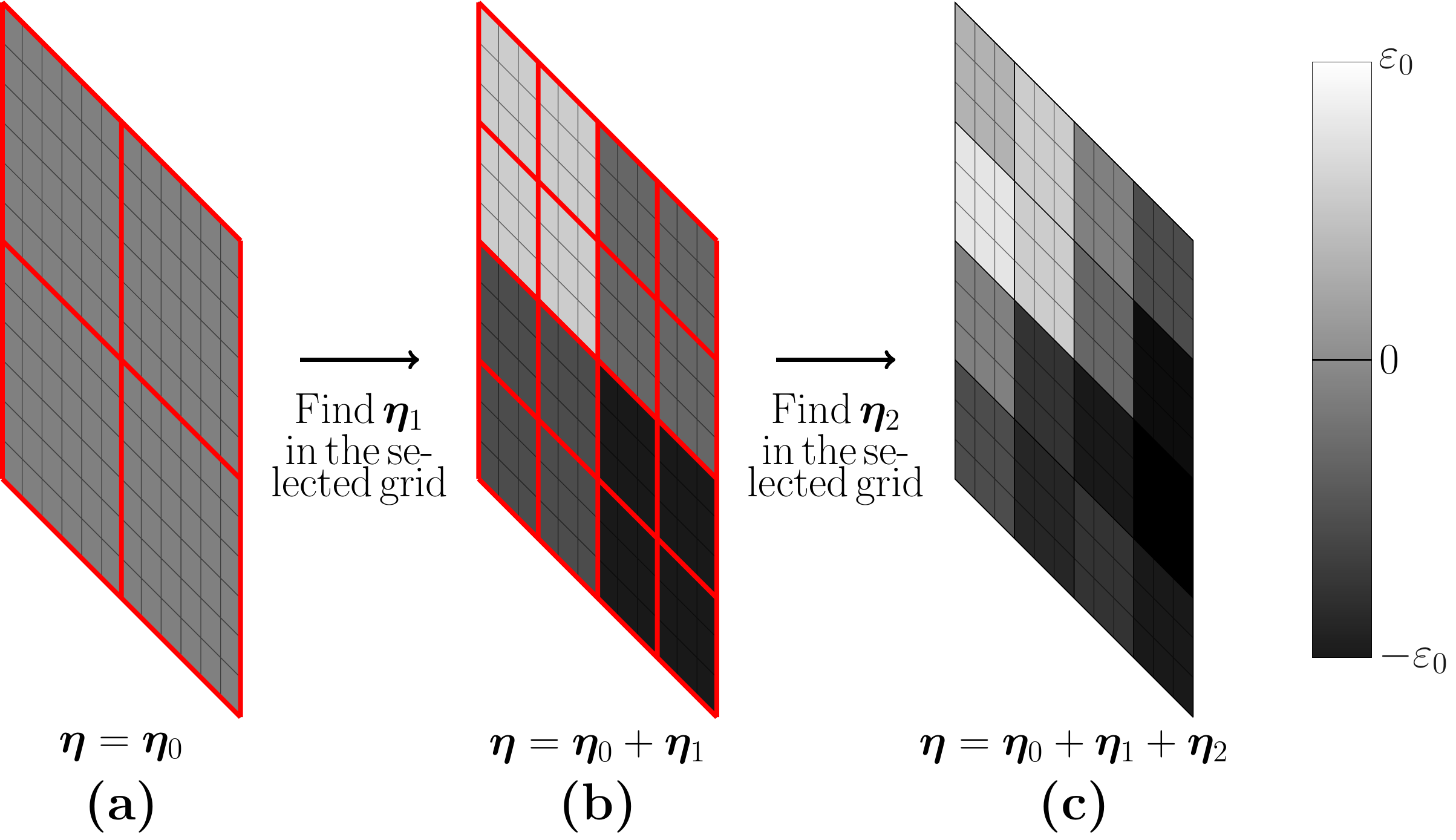}
\caption{Example of how the perturbation $\bm{\eta}$ is generated in a hierarchical lifting method with $m_1=4$ and $m_2=16$ on an image $\in \mathbb{R}^{12\times12}$. In (a) the perturbation is $\bm{\eta} = \bm{\eta}_0$ and we highlight in red the boxes generated via the grid of dimension $m_1$. Once we have found the optimal perturbation $\bm{\eta}_1$, we update the perturbation in (b) and further divide the image with a grid with $m_2$ blocks. Once an optimal solution is found for this grid, the final solution is shown in (c).}
\label{fig:Projections}
\end{figure}

At each level $\ell$ we consider a linear lifting $\textbf{D}^\ell:\mathbb{R}^{m_\ell} \rightarrow \mathbb{R}^n$ and find a level perturbation $\hat{\bm{\eta}}_\ell\in \mathbb{R}^{m_\ell}$ which is added to the full perturbation $\bm{\eta}$, according to
\begin{equation}
    \bm{\eta} = \sum_{j=0}^\ell \bm{\eta}_j= \sum_{j=0}^\ell \textbf{D}^j\hat{\bm{\eta}}_j,
\end{equation}
where $\bm{\eta}_0$ is initialised as $\underline{\textbf{0}}$ and the level perturbations $\bm{\eta}_j$ of the previous layers are considered as fixed. Moreover, we impose that at each level, the grid has to double in refinement, i.e. $m_{\ell+1} = 4m_\ell$. An example of how this works is illustrated in Figure \ref{fig:Projections}. 

\begin{figure}[t]
\centering
\includegraphics[width=.85\linewidth]{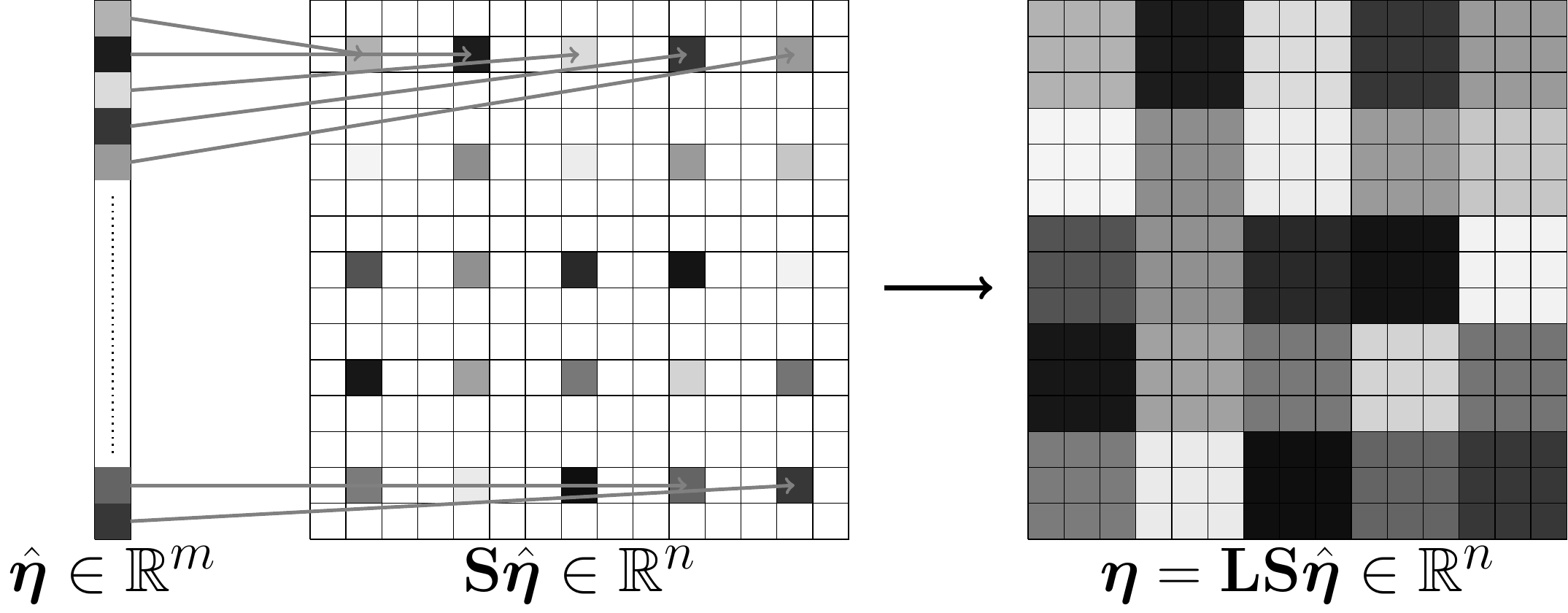}
\caption{In the ``Block'' lifting the perturbation $\bm{\eta}$ is first applier to a sorting matrix $\textbf{S}$ to which an interpolation $\textbf{L}$ is implemented. Thus each block is associated uniquely to one of the variables in $\hat{\bm{\eta}}$.}
\label{fig:Lifting}
\end{figure}

When generating our adversarial examples, we considered two kind of liftings. The first kind of liftings is based on interpolation operations; a sorting matrix  $\textbf{S}^\ell:\mathbb{R}^{m_\ell}\rightarrow\mathbb{R}^n$ is applied such that every index of $\hat{\bm{\eta}}_\ell$ is uniquely associated to a node of a coarse grid masked over the original image. Afterwards, an interpolation $\textbf{L}^\ell:\mathbb{R}^n\rightarrow\mathbb{R}^n$ is implemented over the values in the coarse grid, i.e. $\bm{\eta}_\ell = \textbf{L}^\ell \textbf{S}^\ell\hat{\bm{\eta}}_\ell = \textbf{D}^\ell\hat{\bm{\eta}}_\ell$. The second kind of liftings, instead, forces the perturbation to be high-frequency since there is several literature on these perturbations being the most effective \cite{guo2018,gopalakrishnan2018toward,sharma2019effectiveness}. Some preliminary results lead us to consider the ``Block'' lifting which considers a piecewise constant interpolation and corresponds to the one also used in \cite{COMBI}. Alternative piecewise linear or randomised orderings were also tried, but found not to be appreciably better to justify the added complexity. As we show for the example in Figure \ref{fig:Lifting}, this interpolation lifting divides an image in disjoint blocks via a coarse grid and associates to each of the blocks the same value of a parameter in $\hat{\bm{\eta}}_\ell$. We characterise the lifting $\textbf{D}^\ell$ with the following conditions
\begin{align}
    &\sum_j \textbf{D}_{i,j}^\ell = 1  & \forall i \in \{1,...,n\} \\
    &\sum_i \textbf{D}_{i,j}^\ell = n/m & \forall j \in \{1,...,m\}.
\end{align}

Since $m_\ell$ may still be very high (usually $m_d=n$), for each level $\ell$ we apply domain sub-sampling and consider $\hat{\bm{\eta}}_\ell = \sum_{j=0}^k \hat{\bm{\eta}}_\ell^j$. We order the blocks according to the variance of mean intensity among neighbouring blocks, in contrast to the variance within each block which was suggested in \cite{chen}. Consequently, at each level the adversarial example is found by solving the following iterative problem
\begin{align}
    \min_{\tilde{\bm{\eta}}_\ell^j} \ \ \ & \mathcal{L}\left(\textbf{X} + \tilde{\bm{\eta}},   \textbf{D}^\ell\bm{\Omega}^k\tilde{\bm{\eta}}_\ell^j\right) \;
    s.t.\; \left\|\tilde{\bm{\eta}} +  \textbf{D}^\ell\bm{\Omega}^k\tilde{\bm{\eta}}_\ell^j\right\|_\infty \leq \varepsilon_\infty \label{eq:opt_sub_samp_fin}\\
    & \left[\textbf{X} + \tilde{\bm{\eta}} +  \textbf{D}^\ell\bm{\Omega}^k\tilde{\bm{\eta}}_\ell^j\right]_r \geq l \;\; \forall r\in\{1,...,n\} \nonumber\\
    & \left[\textbf{X} + \tilde{\bm{\eta}} +  \textbf{D}^\ell\bm{\Omega}^k\tilde{\bm{\eta}}_\ell^j\right]_r \leq u \;\; \forall r\in\{1,...,n\},\nonumber \label{eq:opt_prob_proj_end}
\end{align}
where $\tilde{\bm{\eta}} = \sum_{i=0}^{\ell-1}\bm{\eta}_i + \textbf{D}^\ell \sum_{m\neq j} \hat{\bm{\eta}}_\ell^m$.

In its simplest formulation, hierarchical lifting struggles with the pixel-wise interval constraint, $\textbf{X}+\bm{\eta} \in [l,u]^n$. To address this we allow the entries in $\tilde{\bm{\eta}}$ to exceed the interval and then reproject the pixel-wise entries into the interval.

\begin{algorithm}[t]
   \caption{BOBYQA Based Algorithm}
   \label{alg:ALG}
\begin{algorithmic}[1]
   \STATE {\bfseries Input:} Image $\textbf{X}\in \mathbb{R}^n$, target label $t$, maximum perturbation $\varepsilon_\infty$, Neural Net $F$, initial hierarchical level dimensions $m_1$, maximum number of evaluations $n^{max}$, batch sampling size $b$, and maximum number $\kappa$ of queries that we are allowed to do for each batch.
   
   \STATE \textbf{Initialise} $\bm{\eta} \leftarrow \underline{0} \in \mathbb{R}^n$, $n_{eval} = 0$, $\ell=1$.
   
   \WHILE{$\argmax F(\textbf{X}+\bm{\eta})\neq t$ and $n_{eval}<n^{max}$}
   \STATE Compute the number of sub samplings necessary to cover the whole domain $num_{sub}=n/m$
   \STATE Generate the lifting matrix $\textbf{D}_\ell$
   \FOR{$j = 1, \hdots, num_{sub}$}
   \STATE Compute the  matrix $\bm{\Omega}^i$ which selects $b$ dimensions of the $m$-dimensional domain.
   \STATE Define the bounds for a perturbation over the selected pixels of $\textbf{X} + \bm{\eta}$.
   \STATE Find $\hat{\bm{\eta}}_\ell^j$ by implementing the BOBYQA optimisation to the problem (\ref{eq:opt_sub_samp_fin}).
   \STATE Update the noise $\bm{\eta} += \textbf{D}_\ell \hat{\bm{\eta}}_\ell^j$.
   \STATE $n_{eval}$ += $\kappa$, $\ell+=1$, $m*=4$.
   \ENDFOR
   \ENDWHILE
   \IF{$\argmax F(\textbf{X}+\bm{\eta})= t$}
   \STATE The perturbation is successful.
   \ELSIF{$n_{eval} > n^{max}$}
   \STATE The perturbation was not successful with $n^{max}$ iterations.
   \ENDIF
\end{algorithmic}
\end{algorithm}

\subsection{Algorithm pseudo-code}

Our BOBYQA based algorithm is summarised in Algorithm \ref{alg:ALG}; note that not using the hierarchical method corresponds to having one level with $m=n$. A Python implementation of the proposed algorithm based on BOBYQA package from \cite{cartis} is available on Github\footnote{\texttt{https://github.com/giughi/A-Model-Based-}\\\texttt{Derivative-Free-Approach-to-Black-Box}\\\texttt{-Adversarial-Examples-BOBYQA}}.

\section{Comparison of Derivative Free Methods}

We compare the performance of our BOBYQA based algorithm to GenAttack \cite{Alzantot},  combinatorial attacks COMBI \cite{COMBI} and SQUARE \cite{andriushchenko2019square}. The performance is measured by considering the distribution of queries needed to successfully find adversaries to different networks trained on three standard datasets: MNIST \cite{lecun-mnisthandwrittendigit-2010},  CIFAR10 \cite{CIFAR}, and ImageNet \cite{imagenet_cvpr09}.   

\subsection{Parameter Setup for Algorithms}

Our experiments rely for GenAttack \cite{Alzantot}, COMBI \cite{COMBI}, and SQUARE \cite{andriushchenko2019square} on publicly available implementations\footnote{GenAttack: \url{https://github.com/nesl/adversarial_genattack}\\ COMBI: \url{https://github.com/snu-mllab/parsimonious-blackbox-attack} \\ SQUARE: \url{https://github.com/max-andr/square-attack}} with same hyperparameter setting and hierarchical approach as suggested by the respective authors. 

For the proposed algorithm based on BOBYQA, we tuned three main parameters: the dimension of the initial set $q$, the batch dimension $b$, and the trust region radius.

\begin{figure}[t]
\centering
\begin{subfigure}{.22\textwidth}
  \centering
  \includegraphics[width=1\textwidth]{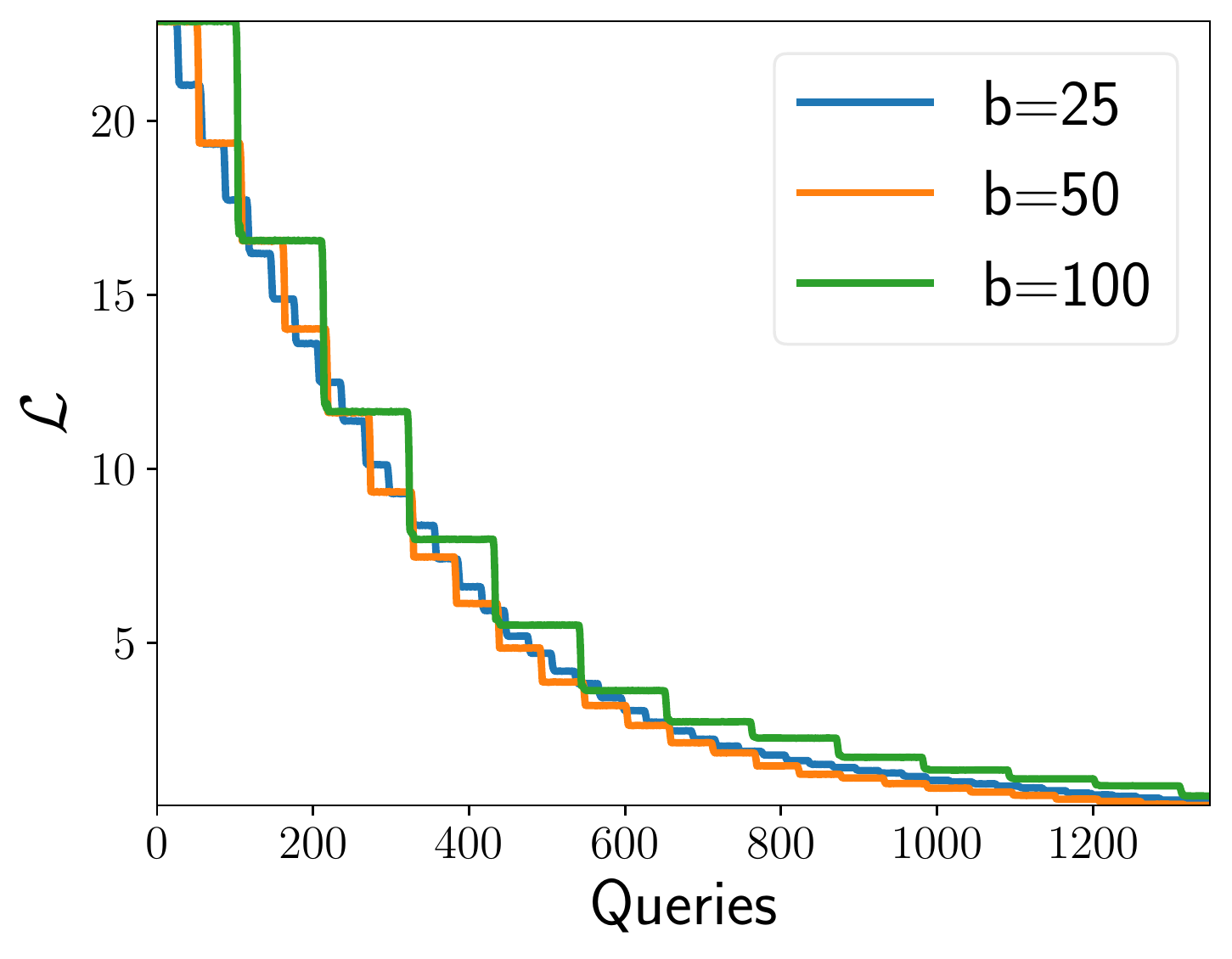}
  \caption{CIFAR10}
  \label{fig:batch_cifar}
\end{subfigure}
\begin{subfigure}{.22\textwidth}
  \centering
  \includegraphics[width=1\textwidth]{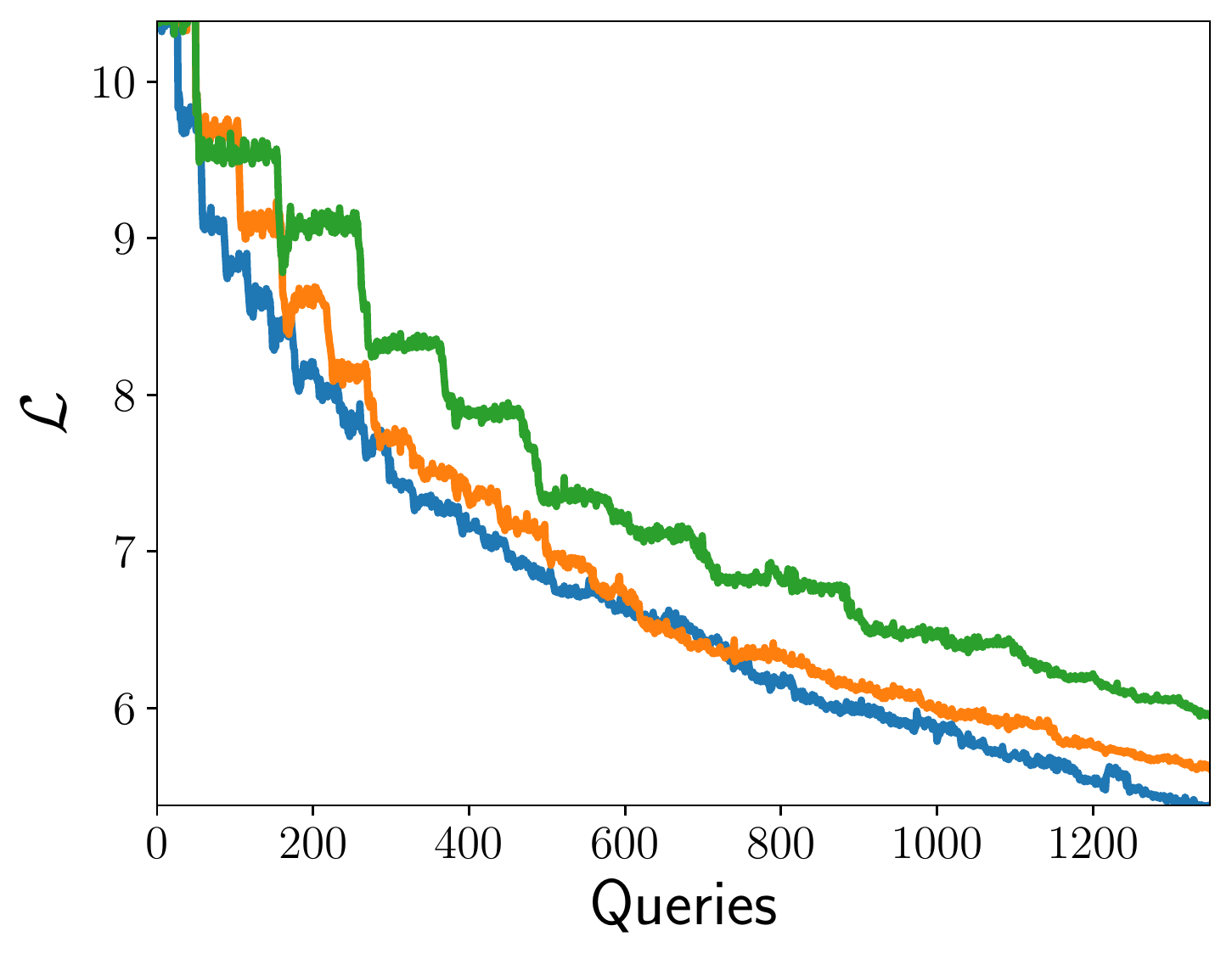}
  \caption{ImageNet}
  \label{fig:batch_Image}
\end{subfigure}%

\caption{Comparison in loss function according to the different batch dimensions $b$ and the different dataset. After the linear model is generated, the optimisation algorithm is always allowed to query the net 5 times if $b=25$ or $b=50$, or 10 times if $b=100$. For ImageNet we are using the hierarchical lifting approach.}
\label{fig:Multi_batch}
\end{figure}

\begin{figure}[t]
\centering
\includegraphics[scale=0.3]{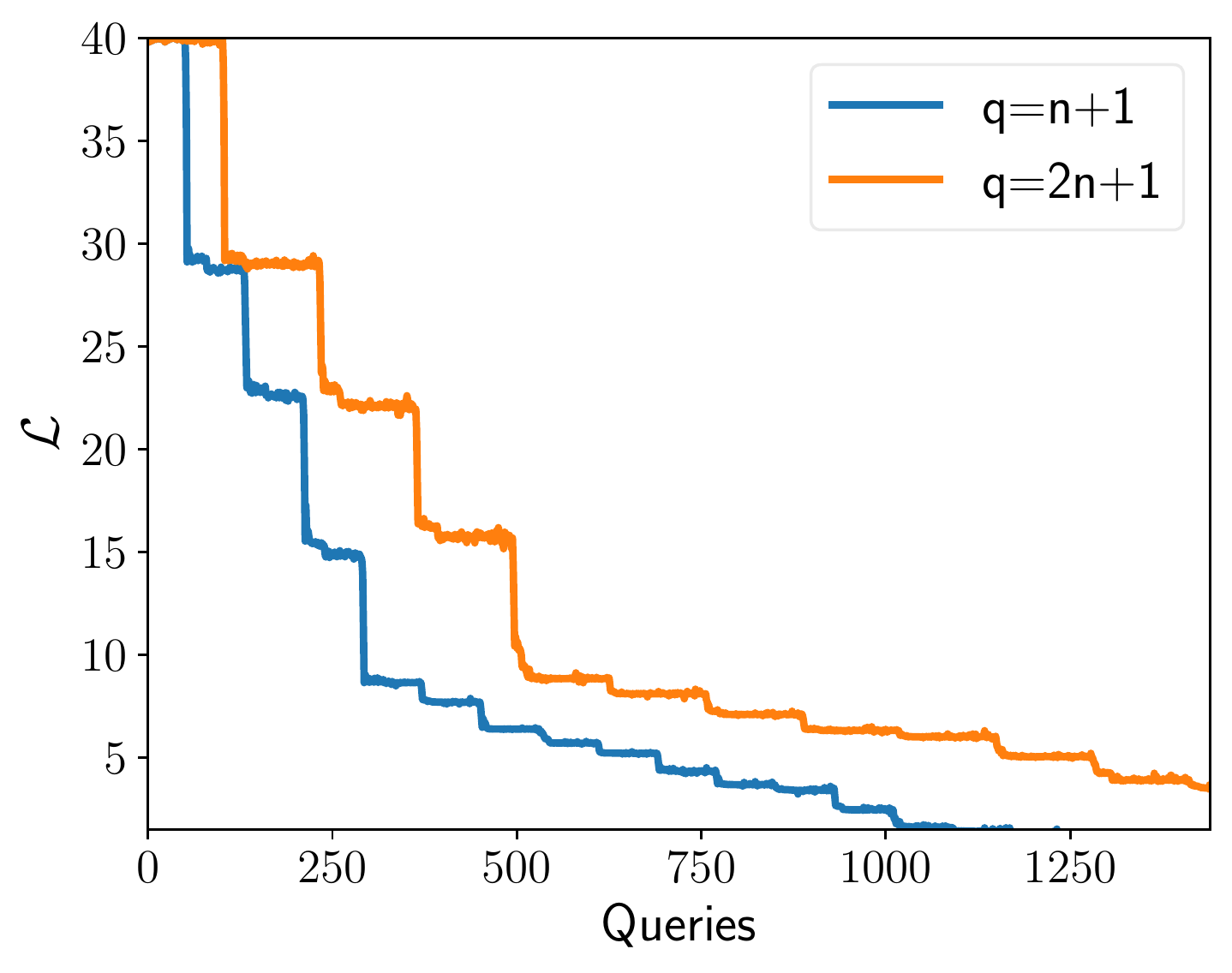}
\caption{Comparison on how the loss $\mathcal{L}$ decreases when the initial set dimension is either $q=b+1$ or $q=2b+1$ in an attack to an image of MNIST with $\varepsilon=0.3$. We chose for both the methods $b=50$ and a maximum of $30$ function evaluations after the model was initialised, i.e. $\kappa = q + 30$.}
\label{fig:Loss_over_queries}
\end{figure}

\begin{description}
    \item{\textbf{Batch Dimension}} Figure \ref{fig:Multi_batch} shows the loss value averaged over 20 images for attacks to NNs trained on CIFAR10, and ImageNet datsets when different batch dimensions are chosen.  The average objective loss as a function of network queries is largely insensitive to the batch sizes, but with modest differences for the larger ImageNet data set where $b=25$ was observed to require modestly fewer queries.  For the remained of the simulations we use  $b=50$ as a good trade-off between faster model generation and good performances.
    
    \item{\textbf{Initial Set Dimension}} Once a subdomain of dimension $b$ is chosen, the model \eqref{eq:Model} is initialised with a set of $q$ samples on which the interpolation conditions \eqref{eq:interpolation} are imposed. There are two main choices for the dimension of the set: either $q=b+1$, thus computing $a$ and $\textbf{c}$ with the interpolation and leaving $\textbf{M}$ always null and thus having a linear model, or $q=2*b+1$ which allows us to initialise $a,\textbf{c}$, and the diagonal of $\textbf{M}$, hence obtaining a quadratic model. The results in Figure \ref{fig:Loss_over_queries} show that at each iteration of the domain sub-sampling the quadratic method performs as well as a linear method, however it requires more queries to initialise the model. Thus we consider the linear model with $q=b+1$\footnote{The Constraint Optimisation by Linear Approximation (COBYLA), a linear based model DFO algorithm, was introduced before BOBYQA \cite{powell2007view}; however, COBYLA considers different constraints on the norm of the variable. Because of this and the possibility to extend the method to quadratic models, we name our algorithm after BOBYQA.}.

    \item{\textbf{Trust Region Radius}} Once the model for the optimisation is built, the step of the optimisation is bounded by the trust region radius. We have selected the beginning radius to be one third of the whole space in which the perturbation lies. With this choice of radius we usually reach within 5 steps a corner of the boundary, and the further iterates remain effectively stationary. 
    
\end{description}

For the hierarchical lifting approach we consider an initial sub-domain of dimension $m_1 = 4 \times 4 \times 3$, as this is the biggest grid that we can optimise over with a batch $b=50$. After considering $m_7 = 256 \times 256 \times 3$, we make use of $m_8 = 299 \times 299 \times 3$ and do not consider further levels.

\subsection{Dataset and Neural Network Specifications}
\label{sec:exp}

Experiments on each dataset are performed with one of the best performing NN architectures as described below

\begin{figure*}[t!]
    \centering
        \rotatebox{90}{\parbox{2mm}{\textbf{MNIST}}}
        \begin{subfigure}{.32\textwidth}
          \centering
          \includegraphics[trim={0.25cm 0.35cm 0.3cm 0.3cm},clip,width=.99\linewidth]{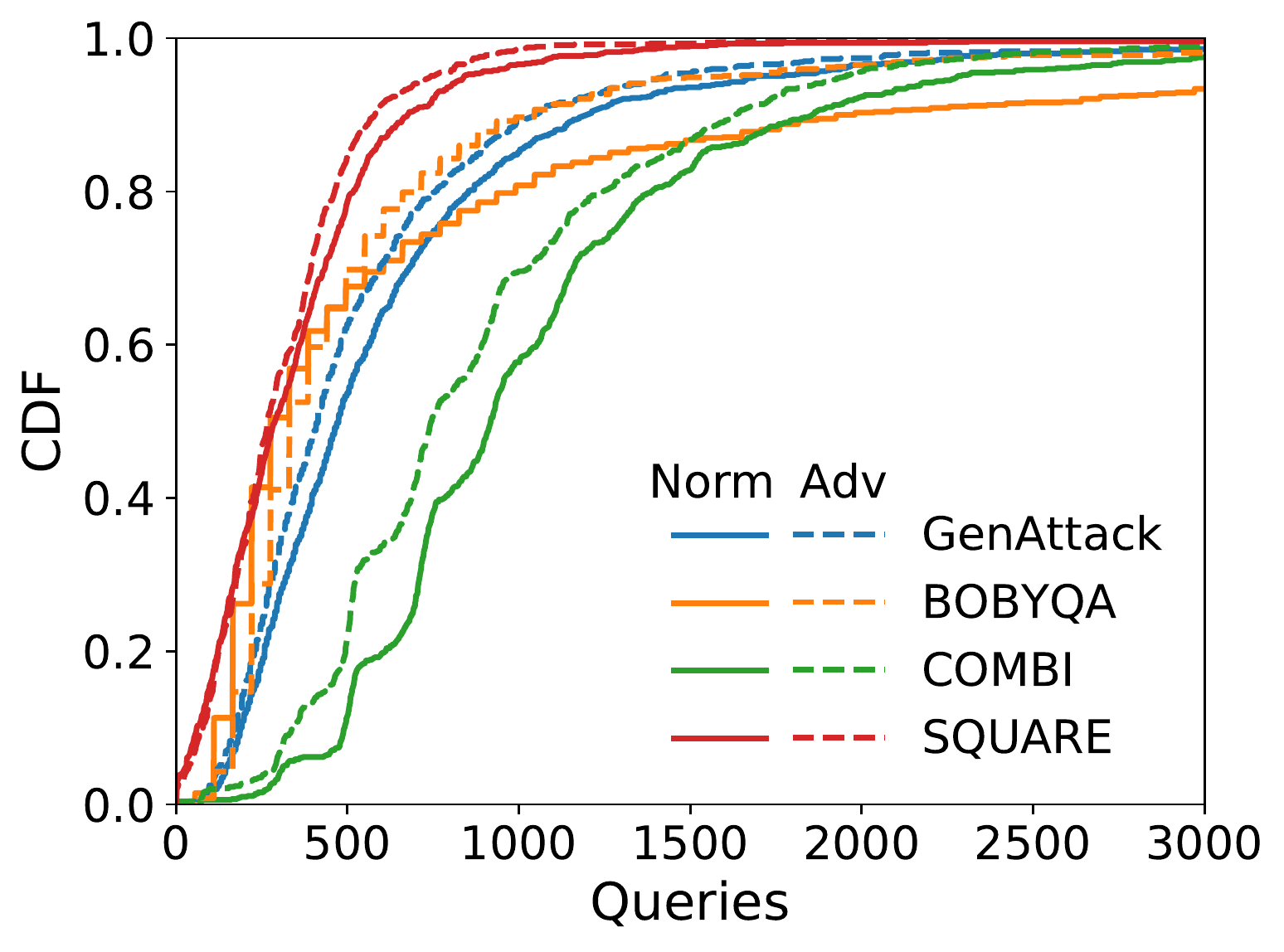}
          \caption{$\varepsilon_\infty=0.4$}
        \end{subfigure}%
        \begin{subfigure}{.32\textwidth}
          \centering
          \includegraphics[trim={0.25cm 0.35cm 0.3cm 0.3cm},clip,width=.99\linewidth]{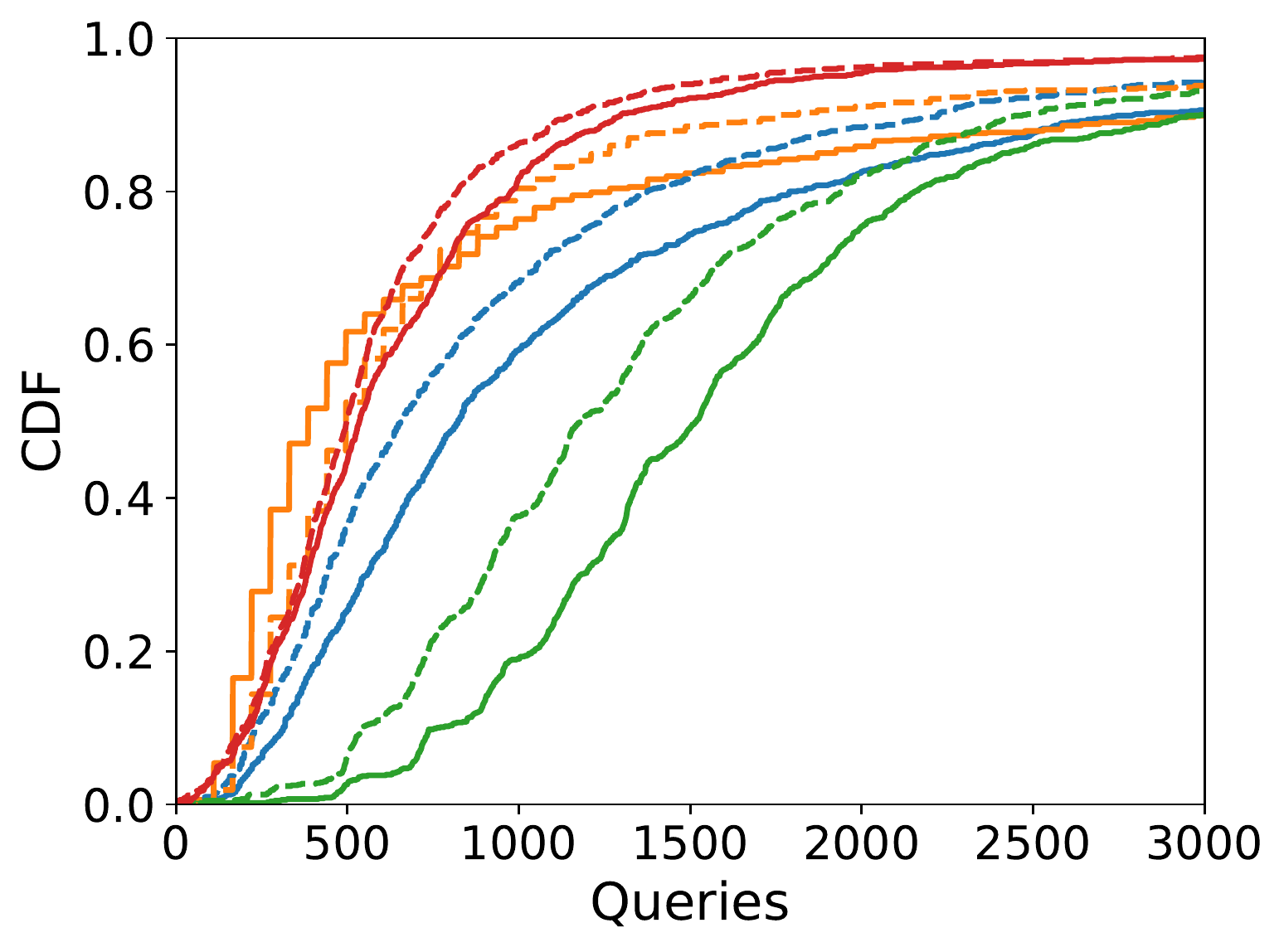}
          \caption{$\varepsilon_\infty=0.3$}
        \end{subfigure}
        \begin{subfigure}{.32\textwidth}
          \centering
          \includegraphics[trim={0.25cm 0.35cm 0.3cm 0.3cm},clip,width=.99\linewidth]{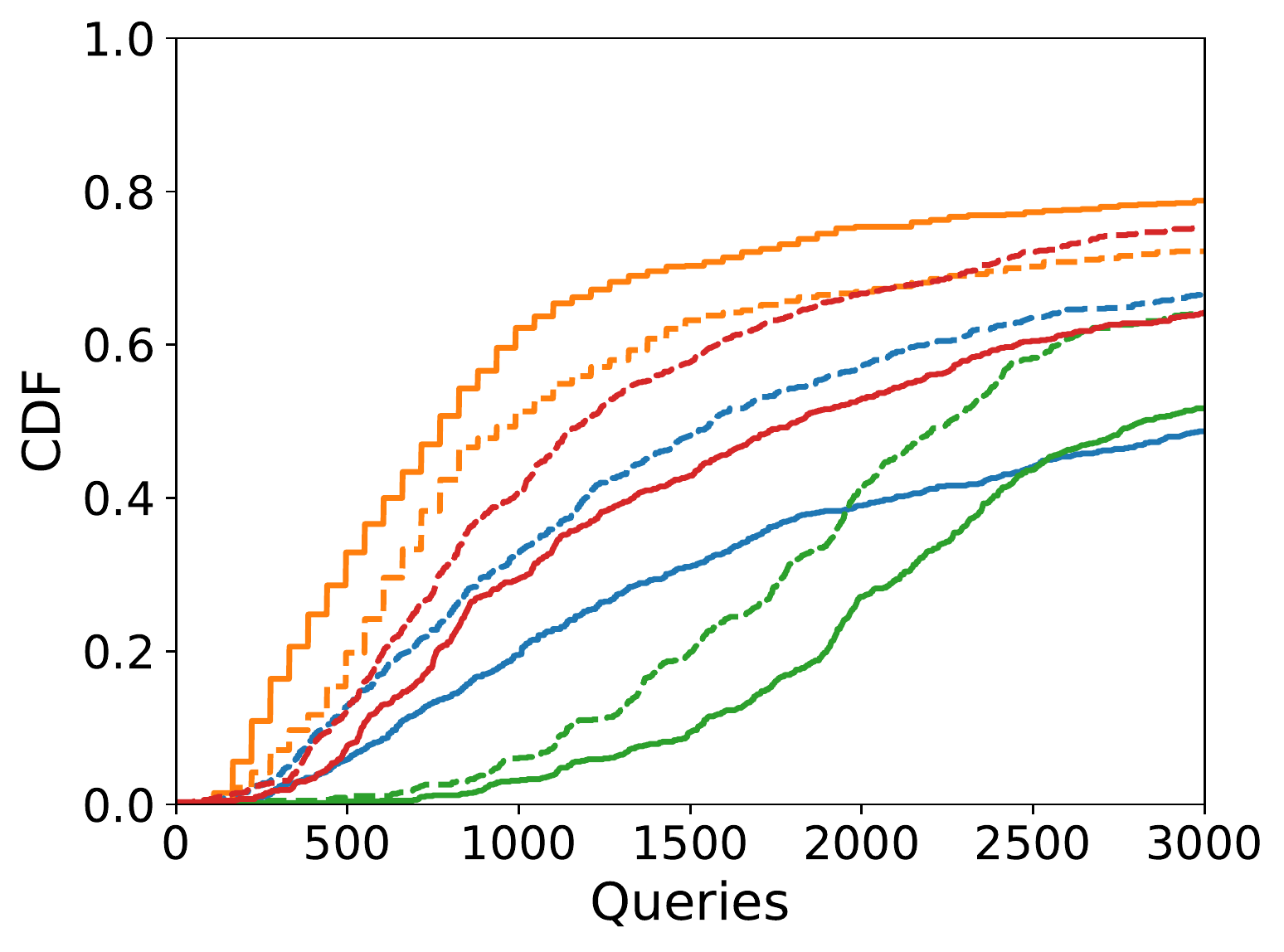}
          \caption{$\varepsilon_\infty=0.2$}
        \end{subfigure}
        \\
        \rotatebox{90}{\parbox{2mm}{\textbf{CIFAR10}}}
        \begin{subfigure}{.32\textwidth}
          \centering
          \includegraphics[trim={0.25cm 0.35cm 0.3cm 0.3cm},clip,width=.99\linewidth]{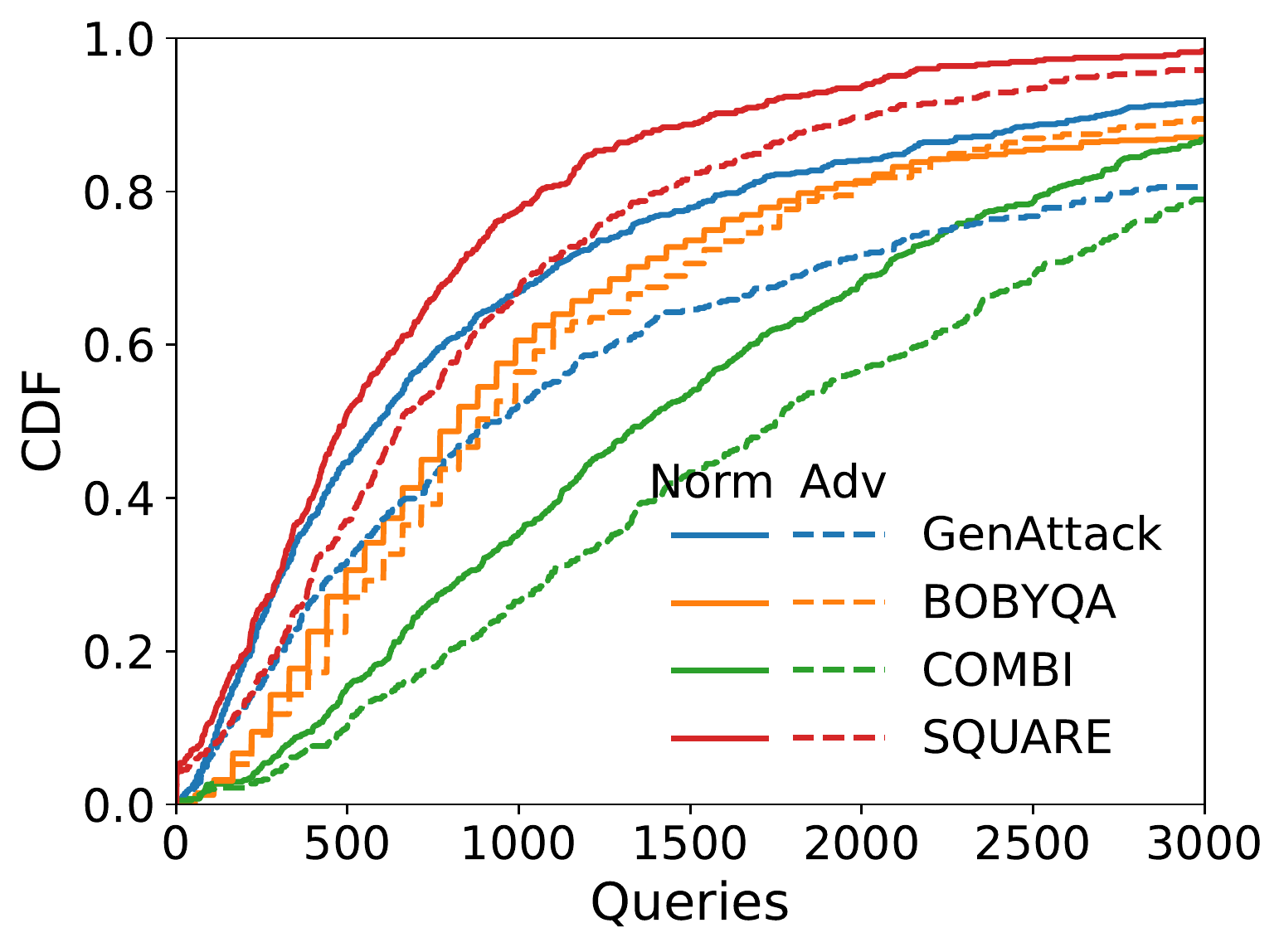}
          \caption{$\varepsilon_\infty=0.1$}
        \end{subfigure}%
        \begin{subfigure}{.32\textwidth}
          \centering
          \includegraphics[trim={0.25cm 0.35cm 0.3cm 0.3cm},clip,width=.99\linewidth]{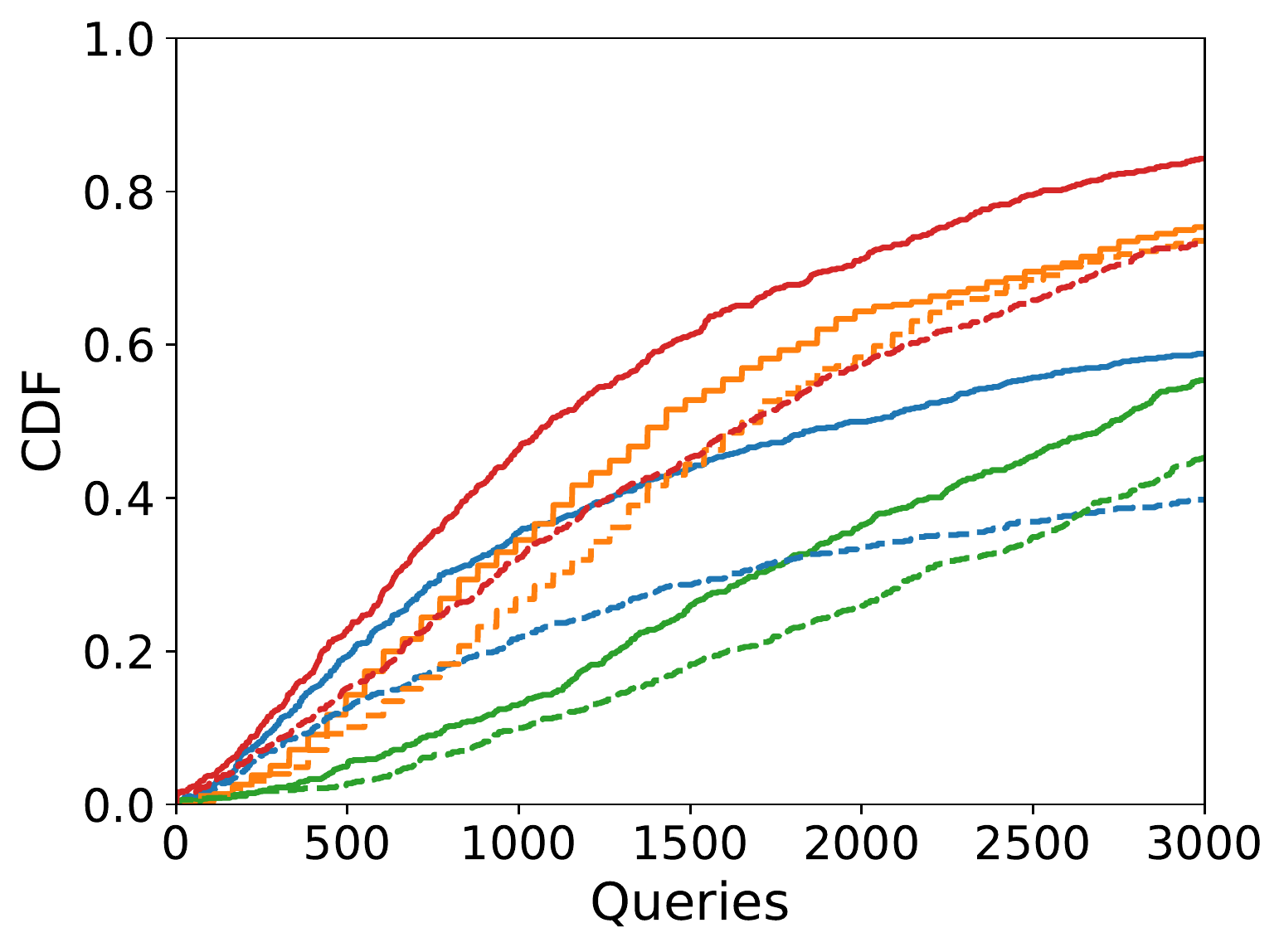}
          \caption{$\varepsilon_\infty=0.05$}
        \end{subfigure}
        \begin{subfigure}{.32\textwidth}
          \centering
          \includegraphics[trim={0.25cm 0.35cm 0.3cm 0.3cm},clip,width=.99\linewidth]{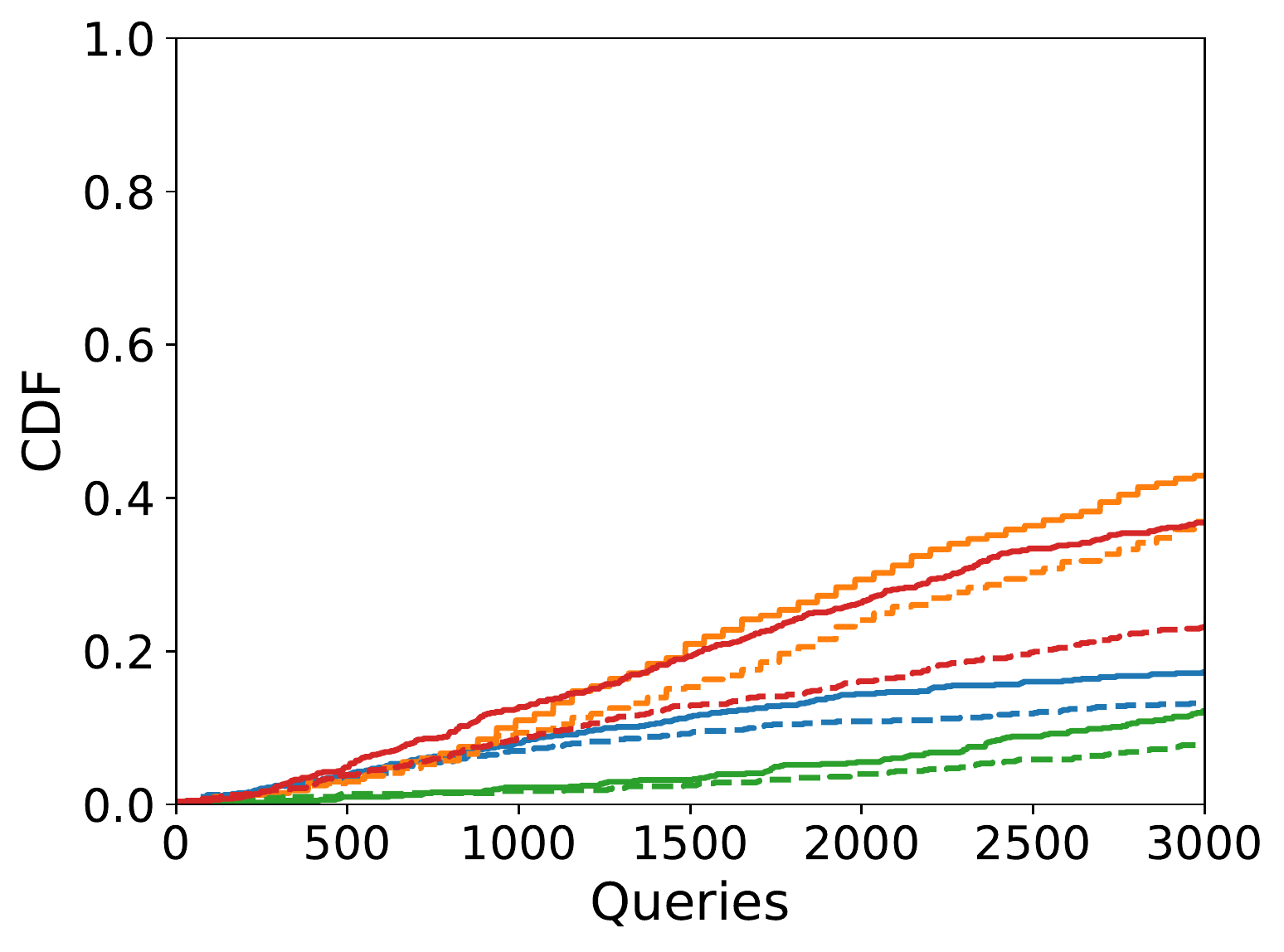}
          \caption{$\varepsilon_\infty=0.02$}
        \end{subfigure}
        \\
        \rotatebox{90}{\parbox{2mm}{\textbf{ImageNet}}}
        \begin{subfigure}{.32\textwidth}
          \centering
          \includegraphics[trim={0.25cm 0.35cm 0.3cm 0.3cm},clip,width=.99\linewidth]{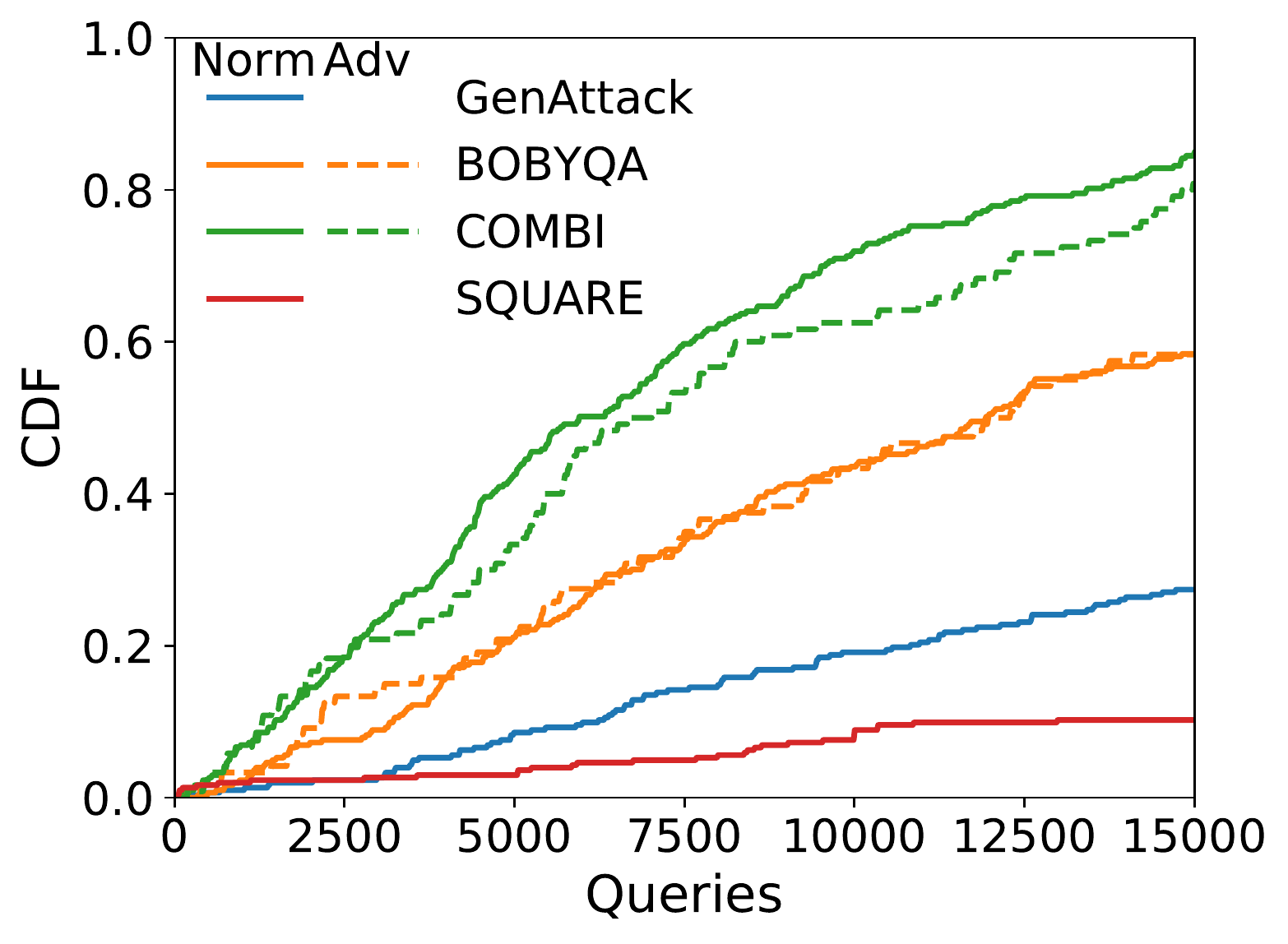}
          \caption{$\varepsilon_\infty=0.05$}
        \end{subfigure}%
        \begin{subfigure}{.32\textwidth}
          \centering
          \includegraphics[trim={0.25cm 0.35cm 0.3cm 0.3cm},clip,width=.99\linewidth]{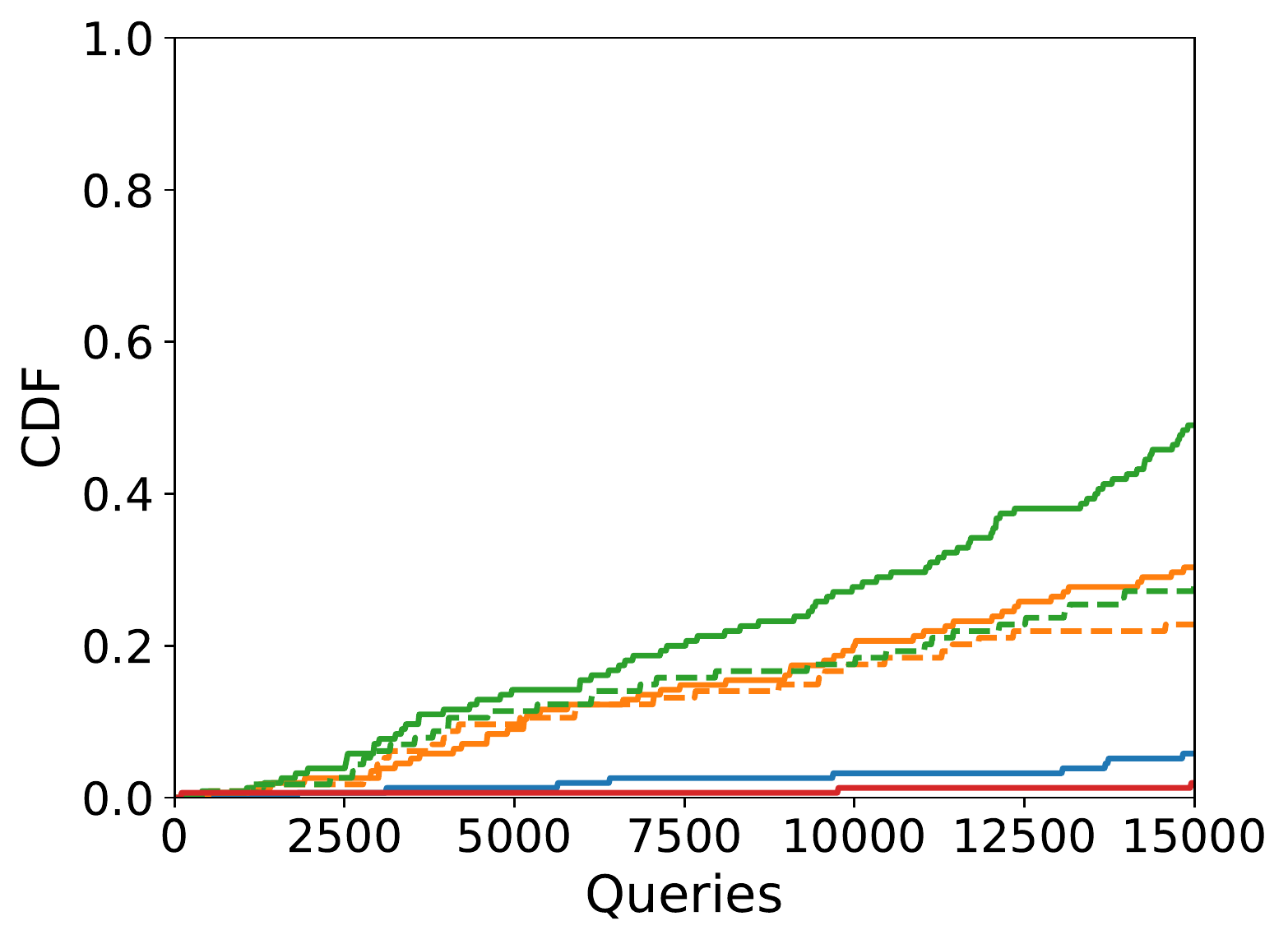}
          \caption{$\varepsilon_\infty=0.02$}
        \end{subfigure}
        \begin{subfigure}{.32\textwidth}
          \centering
          \includegraphics[trim={0.25cm 0.35cm 0.3cm 0.3cm},clip,width=.99\linewidth]{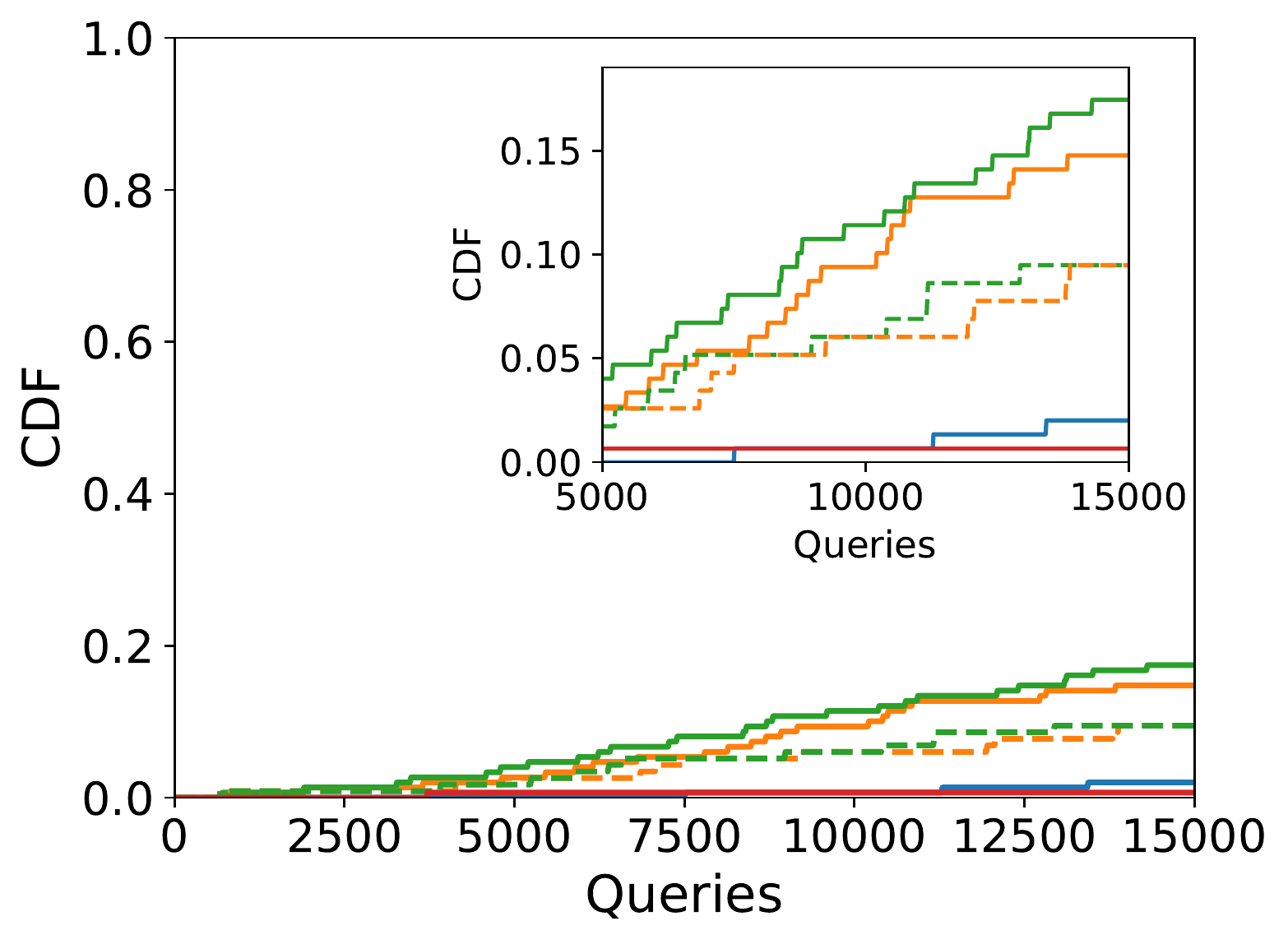}
          \caption{$\varepsilon_\infty=0.01$}
        \end{subfigure}
        \caption{Cumulative fraction of test set images successfully misclassified with adversarial examples generated by GenAttack, COMBI, SQUARE and our BOBYQA based approach for different perturbation energies $\varepsilon_\infty$ and NNs trained on MNIST, CIFAR10 and ImageNet dataset.
        In all results the solid and dashed lines denoted by `Norm' and `Adv' corresponds to attacks on nets trained without or with a defence strategy respectively. For MNIST and CIFAR we consider the distillation defence method from \cite{Papernot_distillation} while for ImageNet the adversarial training proposed in \cite{kurakin2016adversarial}.
        }
        \label{fig:CIFARcdf}
\end{figure*}

\paragraph{MNIST/CIFAR10} 

MNIST and CIFAR10 are two data-sets with images divided between 10 classes and of dimension 28x28x1 and 32x32x3 respectively. On them we apply the net introduced in \cite{chen} which is structured in succession by: 2 Conv layers with ReLu activation followed by a maxpooling layer. This process is repeated twice and then two dense layers with Relu activation are applied. Finally a softmax layer generates the output vector. For each dataset, we train the same architecture in two different ways obtaining separate nets. One is obtained by optimising the accuracy of the net on raw unperturbed images, while the other is trained with the application of the distillation defence by \cite{Papernot_distillation}.

To generate a comprehensive distribution for the queries at each energy budget, for both the two trained nets and 10 images per class, we attempt to misclassify an image targeting all of the 9 remaining classes; this way we generate a total of 900 perturbations per energy budget. For these two datasets the images are of relative low dimension and we do not apply the hierarchical approach.

\paragraph{ImageNet} This is a data-set of millions of images with a dimension of 299x299x3 divided between 1000 classes. For this data-set we consider the Inception-v3 net \cite{Inceptionv3} trained with and without the adversarial defence proposed in \cite{kurakin2016adversarial}\footnote{For the non-adversarially trained net we considered the one available at \url{http://jaina.cs.ucdavis.edu/datasets/adv/imagenet/inception_v3_2016_08_28_frozen.tar.gz}, while for the weights of the adversarially trained net we relied on \url{https://github.com/tensorflow/models/tree/master/research/adv_imagenet_models}.}. Due to the large number of target classes in ImageNet, we perform tests on random images and target classes.  The number of tests conducted for Inception-v3 \cite{Inceptionv3} and the adversarially trained variant \cite{kurakin2016adversarial} are: 303 and 120 for $\epsilon_{\infty}=0.05$, 155 and 114 for $\epsilon_{\infty}=0.02$ and 149 and 116 for $\epsilon_{\infty}=0.01$ respectively.

\subsection{Experimental Results}

In Figure \ref{fig:CIFARcdf} we present the cumulative fraction of images misclassified (abridged by CDF for cumulative distribution function)  as a function of the number of queries to the NN for different perturbation energies $\varepsilon_\infty$. The pixels are normalised to be in the interval $(-1/2,1/2)$, hence, $\varepsilon_\infty=0.1$ would imply that any pixel is allowed to change $10\%$ of the total intensity range from its initial value. By illustrating the CDFs we easily see which method has been able to misclassify the largest fraction of images in the given test-set for a fixed number of queries to the NN.
It can be observed that the proposed BOBYQA based approach achieves state-of-the-art results
when the perturbation bound of $\bm{\eta}$ decreases. This behaviour is consistent across all of the considered datasets (MNIST, CIFAR10, and ImageNet); however, the energy at which the BOBYQA algorithm performs the best, varies in each case. 

In the experiments we also considered nets trained with defence methods, distillation \cite{Papernot_distillation} for MNIST and CIFAR10 datasets while adversarial training \cite{kurakin2016adversarial} for ImageNet, and the results can be identified in Figure \ref{fig:CIFARcdf} by the dashed lines. Similar to the previous case, we observe that the proposed BOBYQA based algorithm performs the best when the energy perturbation decreases. Moreover, the BOBYQA based algorithm seems to be the least affected in its performance when the any defence is used; for example, at $\epsilon_{\infty}=$0.01 and 15,000 queries, the defence reduces the CDF of COMBI by 0.078 compared to 0.051 for BOBYQA. This further supports the idea that for more challenging scenarios model-based approaches are preferable as compared to model-free counterparts. 

We associate the counter-intuitive improvement of the CDF in the MNIST and ImageNet  with high perturbation energies cases to the distillation and the adversarial training being focused primarily on low energy perturbations. For ImageNet, non-model-based algorithms use different hierarchical approaches which we expect leads in part to the  superior performance of COMBI in Fig.~\ref{fig:CIFARcdf} panels (g)-(i). 


\section{Discussion and Conclusion}

We have introduced BOBYQA, a method to search adversarial examples based on a model-based DFO algorithm and have conducted some experiments to understand how it compares to existing GenAttack \cite{Alzantot}, COMBI \cite{COMBI}, and SQUARE \cite{andriushchenko2019square} attack, when targeted black-box adversarial examples are searched with the fewest queries to a neural net. 

Following the results of the experiments that we presented above, the method with which generating the adversarial example should be chosen according to which setting the adversary is considering. When the perturbation energy is high, one should choose either COMBI if the input is high-dimensional or SQUARE if the input is low-dimensional. On the other hand,  a model-based approach like BOBYQA should be considered as soon as the complexity of the setting increases, e.g. the maximum perturbation energy is reduced or the net is adversarially trained.

With the BOBYQA attack algorithm we have introduced a different approach for the generation of targeted adversarial examples in a black-box setting with the aim of exploring what advantages are achieved by considering model-based DFO algorithms. We did not focus on presenting an algorithm which is in absolute the most efficient; primarily because our algorithm has several aspects in which to be improved.
The BOBYQA attack is limited by the implementation of py-BOBYQA \cite{cartis} since the element-wise constraints do not allow the consideration of more sophisticated liftings which leverage on compressed sensing, to name one of the many possible variations. 

In conclusion, the results in this paper support how sophisticated misclassification methods are preferable in challenging settings. As a consequence, variations on our model-based algorithms should be considered in the future as a tool to establish the effectiveness of newly presented adversarial defence techniques.

\section*{Acknowledgements}

This publication is based on work supported by the EPSRC Centre for Doctoral Training in Industrially Focused Mathematical Modelling (EP/L015803/1) in collaboration with New Rock Capital Management. 

\bibliography{bibliography}
\bibliographystyle{icml2020}



\end{document}